\documentclass[twoside]{article}

\usepackage[accepted]{aistats2020}

\usepackage[utf8]{inputenc} 
\usepackage[T1]{fontenc}    
\usepackage{microtype}      

\usepackage{amsmath}
\usepackage{amssymb}
\usepackage{amsfonts}       
\usepackage{nicefrac}       
\usepackage{mathtools}
\usepackage{bm}
\usepackage{physics}
\usepackage{algorithm}
\usepackage{algpseudocode}

\usepackage{amsthm}
\newtheoremstyle{theorem-style}
  {\topsep} 
  {\topsep} 
  {\itshape} 
  {} 
  {\bfseries} 
  {} 
  {\newline} 
  {} 
\theoremstyle{theorem-style}
\newtheorem{theorem}{Theorem}

\newtheoremstyle{definition-style}
  {\topsep} 
  {\topsep} 
  {} 
  {} 
  {\bfseries} 
  {} 
  {\newline} 
  {} 
\theoremstyle{definition-style}

\usepackage{booktabs}       
\usepackage{csvsimple}
\usepackage{graphicx}
\usepackage{subcaption}
\usepackage[figuresright]{rotating}

\usepackage[round]{natbib}

\usepackage{url}            
\setcitestyle{comma, numbers, sort&compress}
\usepackage{hyperref}
\usepackage{cleveref}
\usepackage{bookmark} 

\usepackage[prependcaption,textsize=small,color=gray!40, disable]{todonotes} 

\newcommand{\E}[1]{\mathbb{E}\left[#1\right]}

\usepackage{amsmath,amsfonts,bm}

\def\1{\bm{1}}

\def\ry{{\textnormal{y}}}
\def\rz{{\textnormal{z}}}

\def\rvepsilon{{\mathbf{\epsilon}}}

\def\rvg{{\mathbf{g}}}

\def\rvu{{\mathbf{i}}}

\def\rvu{{\mathbf{u}}}

\def\rvw{{\mathbf{w}}}
\def\rvx{{\mathbf{x}}}
\def\rvy{{\mathbf{y}}}
\def\rvz{{\mathbf{z}}}

\def\ervg{{\textnormal{g}}}

\def\ervz{{\textnormal{z}}}

\def\vmu{{\bm{\mu}}}
\def\vtheta{{\bm{\theta}}}
\def\vphi{{\bm{\phi}}}

\def\vm{{\bm{m}}}

\def\vx{{\bm{x}}}

\def\mA{{\bm{A}}}

\def\mC{{\bm{C}}}

\def\mM{{\bm{M}}}

\def\mS{{\bm{S}}}

\def\mSigma{{\bm{\Sigma}}}

\DeclareMathAlphabet{\mathsfit}{\encodingdefault}{\sfdefault}{m}{sl}
\SetMathAlphabet{\mathsfit}{bold}{\encodingdefault}{\sfdefault}{bx}{n}

\def\gM{{\mathcal{M}}}

\newcommand{\R}{\mathbb{R}}

\DeclareMathOperator*{\argmax}{arg\,max}

\newcommand{\papertitle}{Non-Parametric Calibration for Classification}

\newcommand{\firstauthor}{Jonathan Wenger}
\newcommand{\secondauthor}{Hedvig Kjellstr\" om}
\newcommand{\thirdauthor}{Rudolph Triebel}

\newcommand{\firstaddress}{University of Tuebingen\\ TU Munich\\KTH Royal Institute of Technology\\\texttt{jonathan.wenger@uni-tuebingen.de}}
\newcommand{\secondaddress}{KTH Royal Institute of Technology\\ \\~\\ \texttt{hedvig@kth.se}}
\newcommand{\thirdaddress}{German Aerospace Center (DLR)\\ TU Munich\\ \\ \texttt{rudolph.triebel@dlr.de}}

\begin{document}

%

%

\twocolumn[

\aistatstitle{\papertitle{}}

\ifnum\statePaper=1
\aistatsauthor{\firstauthor{} \And \secondauthor{} \And \thirdauthor{}}
\aistatsaddress{ \firstaddress{} \And  \secondaddress{} \And \thirdaddress{} }

\else

\aistatsauthor{ Author 1 \And Author 2 \And  Author 3 }
\aistatsaddress{ Institution 1 \And  Institution 2 \And Institution 3 }

\fi]

\begin{abstract}
Many applications of classification methods not only require high accuracy but also reliable estimation of predictive uncertainty. However, while many current classification frameworks, in particular deep neural networks, achieve high accuracy, they tend to incorrectly estimate uncertainty. In this paper, we propose a method that adjusts the confidence estimates of a general classifier such that they approach the probability of classifying correctly. In contrast to existing approaches, our calibration method employs a non-parametric representation using a latent Gaussian process, and is specifically designed for multi-class classification. It can be applied to any classifier that outputs confidence estimates and is not limited to neural networks. We also provide a theoretical analysis regarding the over- and underconfidence of a classifier and its relationship to calibration, as well as an empirical outlook for calibrated active learning. In experiments we show the universally strong performance of our method across different classifiers and benchmark data sets, in particular for state-of-the art neural network architectures.
\end{abstract}

\section{INTRODUCTION}
With the recent achievements in machine learning, in particular in the area of deep learning, the application range for learning methods has increased significantly. Especially in challenging fields such as computer vision or speech recognition, important advancements have been made using powerful and complex network architectures, trained on very large data sets. Most of these techniques are used for classification tasks, e.g. object recognition. We also consider classification in our work. However, in addition to achieving high classification accuracy, our goal is to provide reliable prediction uncertainty estimates. This is particularly relevant in safety-critical applications, such as autonomous driving and robotics \citep{Amodei2016}. Reliable uncertainties can be used to increase a classifier's precision by reporting only class labels that are predicted with low uncertainty or for information theoretic analyses of what was learned and what was not. The latter is especially interesting in active learning, where the model actively selects the most relevant data samples for training via a query function based on the predictive uncertainty of the model \citep{Settles2010}.
 
\begin{figure*}
\centering
\includegraphics[width=0.9\textwidth]{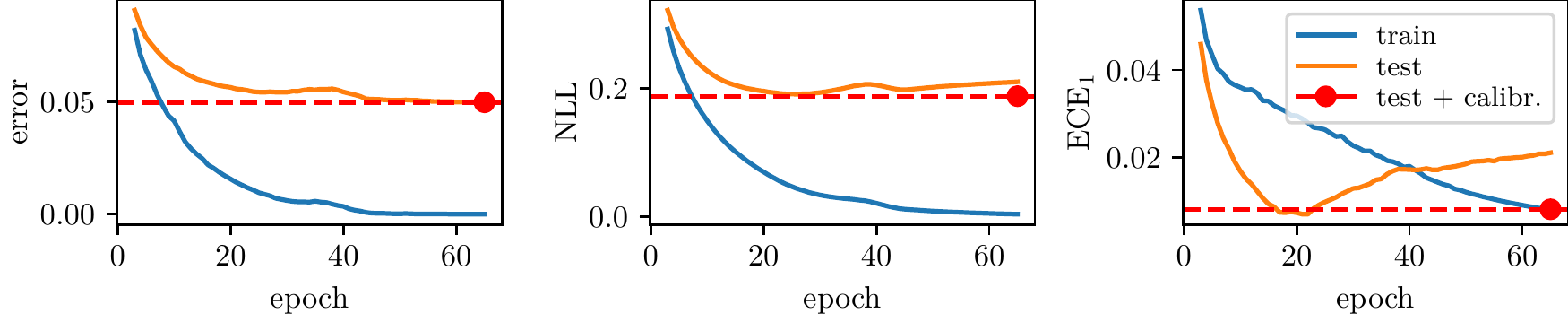}
\caption{\textit{Motivating example for calibration.} We trained a neural network with one hidden layer on MNIST \citep{LeCun1998} and computed the classification error, the negative log-likelihood (NLL) and the expected calibration error (\(\text{ECE}_1\)) for each training epoch. While accuracy continues to improve on the test set, the \(\text{ECE}_1\) increases after 20 epochs. This differs from classical overfitting as the test error continues to decrease. This indicates that improving both accuracy and uncertainty estimation can be conflicting objectives. However, we can mitigate this post-hoc via our calibration method (red dot). The uncertainty estimation after training and calibration is improved with maintained classification accuracy. \label{fig:illustration_ece_logloss}}
\end{figure*}

Unfortunately, current probabilistic classification approaches that inherently provide good uncertainty estimates, such as Gaussian processes (GP), often suffer from lower accuracy and higher computational complexity on high-dimensional classification tasks compared to state-of-the-art convolutional neural networks (CNN). It was recently observed that many modern CNNs are overconfident \citep{Lakshminarayanan2017, Hein2019} and miscalibrated \citep{Guo2017}. Calibration refers to how well  confidence estimates of a classifier match the probability of the associated prediction being correct. Originally developed in the context of forecasting \citep{Murphy1973, DeGroot1983}, uncertainty calibration has seen an increased interest in recent years \citep{Naeini2015, Guo2017, Vaicenavicius2019}, partly because of the popularity of CNNs which generally lack an inherent uncertainty representation. Earlier studies show that also classical methods such as decision trees, boosting, SVMs and naive Bayes classifiers tend to be miscalibrated \citep{Zadrozny2001, Niculescu-Mizil2005, Niculescu-Mizil2005a, Naeini2015}. Based on these observations, we claim that training and calibrating a classifier can be two different objectives that benefit from being considered separately, as shown in a toy example in \Cref{fig:illustration_ece_logloss}. Here, a simple neural network continually improves its accuracy on the test set during training, but eventually overfits in terms of NLL and calibration error. A similar phenomenon was observed by \citet{Guo2017} for more complex models.

Calibration methods approach this problem by performing a post-hoc improvement to uncertainty estimation using a small subset of the training data. Our goal in this paper is to develop a \emph{multi-class calibration method for arbitrary classifiers}, to provide reliable predictive uncertainty estimates in addition to maintaining high accuracy. In contrast to recent approaches, which strive to improve uncertainty estimation only for neural networks, including Bayesian neural networks \citep{MacKay1992,Gal2016} and Laplace approximations (LA) \citep{Martens2015,Ba2017}, our aim is a framework that is not based on tuning a specific classification method. This has the advantage that our method operates independently of the training process.

\paragraph{Contribution} 
In this work we develop a new multi-class and model-agnostic approach to calibration, based on a latent Gaussian process inferred using variational inference. We replicate and extend previous findings that popular classification models are generally not calibrated and demonstrate the superior performance of our method for deep neural networks. Finally, we study the relationship between active learning and calibration from a theoretical perspective and give an empirical outlook.

\paragraph{Related Work}
Estimation of uncertainty, in particular in deep learning \citep{Kendall2017}, is of considerable interest in the machine learning community. There are two main approaches in classification. The first chooses a model and a (regularized) loss function for a particular problem to inherently learn a good representation, and the second performs post-hoc calibration by transforming the output of the underlying model. For example, \cite{Pereyra2017} propose to penalize low-entropy output distributions, \cite{Kumar2018} suggest a trainable measure of calibration as a regularizer and \cite{Maddox2019} employ an approximate Bayesian inference technique using stochastic weight averaging. \cite{Milios2018} approximate Gaussian process classifiers by GP regression on transformed labels for better scalability and \cite{Wilson2016} combine additive Gaussian processes with deep neural network architectures.
Research on calibration goes back to statistical forecasting \citep{Murphy1973,DeGroot1983} and approaches to provide uncertainty estimates for non-probabilistic binary classifiers \citep{Platt1999, Lin2007, Zadrozny2002}. More recently, Bayesian binning into quantiles \citep{Naeini2015} and beta calibration \citep{Kull2017a} for binary classification and temperature scaling \citep{Guo2017} for multi-class problems were proposed. A theoretical framework for evaluating calibration in classification was suggested by \cite{Vaicenavicius2019}. Calibration was also previously considered in the online setting with potentially adversarial input \citep{Kuleshov2017}.
Calibration in a broader sense is also of interest outside of the classification setting, e.g. in regression \citep{Kuleshov2018,Song2019}, in the discovery of causal Bayesian network structure from observational data \citep{Jabbari2017} and in the algorithmic fairness literature \citep{Pleiss2017, Kleinberg2018}.

\section{BACKGROUND}
\label{sec:background}

\paragraph{Notation}
Consider a data set \(\mathcal{D}=\left \{(\vx_{n}, y_{n})\right \}_{n=1}^N\) assumed to consist of independent and identically distributed realizations of the random variable \((\rvx,\ry) \in \mathcal{X} \times \mathcal{Y}\) with \(K \coloneqq\abs{\mathcal{Y}}\) classes. If not stated otherwise, any expectation is taken with respect to the law of \((\rvx,\ry)\). Let \(f : \mathcal{X} \rightarrow \mathbb{R}^K\) be a classifier with output \(\rvz = f(\rvx)\), prediction \(\hat{\ry} = \argmax_{i}(\ervz_i)\) and associated confidence score \(\hat{\rz} = \max_{i}(\ervz_i)\). Lastly, \(v : \mathbb{R}^K \rightarrow \mathbb{R}^K\) denotes a calibration method.

\subsection{Calibration}
\label{sec:calibration}
A classifier is called \textit{calibrated} if the confidence in its class prediction matches the probability of its prediction being correct, i.e. \(\mathbb{E}\left[1_{\hat{\ry} = \ry} \mid \hat{\rz} \right] = \hat{\rz}\). In order to measure calibration, we define the \textit{expected calibration error} following \cite{Naeini2015} for \(1 \leq p < \infty\) by
\begin{equation}
\label{eqn:exp_calibration_error}
\textup{ECE}_p = \mathbb{E} \big[\abs{\hat{\rz} - \mathbb{E}\left[1_{\hat{\ry} = \ry} \mid \hat{\rz}\right]}^p\big]^{\frac{1}{p}}
\end{equation}
and the \textit{maximum calibration error} by \(\textup{ECE}_{\infty} = \max_{z \in [0,1]} \abs{\hat{\rz} - \mathbb{E}\left[1_{\hat{\ry} = \ry} \mid \hat{\rz} = z \right]}\). In practice, we estimate the calibration error using a fixed binning for \(\hat{\rz}\) as described by \cite{Naeini2015}. However, calibration alone is not sufficient for useful uncertainty estimates. A classifier on a balanced binary classification problem that always returns a confidence of $0.5$ is perfectly calibrated, because this equals the probability of making a correct prediction. However, intuitively prediction confidence should be sufficiently close to 0 and 1 to be informative. This notion is known as sharpness or refinement \citep{DeGroot1983,Murphy1992,Cohen2004}.

\subsection{Over- and Underconfidence}
\label{sec:over_underconfidence}
We build on the notions of \emph{over-} and \emph{underconfidence}, as introduced previously in the context of active learning by \cite{Mund2015}. The idea is to measure the average confidence of a classifier on its false predictions and the average uncertainty on its correct predictions:
\begin{equation}
\label{eqn:over_underconfidence}
o(f) = \mathbb{E}\left[\hat{\rz} \mid \hat{\ry} \neq \ry \right] \quad u(f) = \mathbb{E}\left[1-\hat{\rz} \mid \hat{\ry} = \ry\right]
\end{equation}
Over- and underconfidence are properties of a classifier independent of accuracy. They relate to query efficiency in active learning \citep{Settles2010}. Counter to intuition, both can be present to varying degrees simultaneously. We refer to \Cref{sec:supp_over_underconfidence} of the supplementary material for more details. We demonstrate that there is a direct link between calibration and these two notions.

\begin{theorem}[Calibration, Over- and Underconfidence]
\label{thm:calibration_overconfidence}
Let \(1 \leq p < q \leq \infty\), then the following relationship between overconfidence, underconfidence and the expected calibration error holds:
\begin{equation*}
\abs{o(f) \mathbb{P}(\hat{\ry} \neq \ry) -u(f) \mathbb{P}(\hat{\ry} = \ry)} \leq \textup{ECE}_p \leq \textup{ECE}_q.
\end{equation*}
\end{theorem}
A proof is given in \Cref{sec:supp_proofs} of the supplementary material. We see that the expected calibration error bounds the weighted absolute difference of over- and underconfidence. This implies that for perfect calibration, the odds of making a correct prediction equal the ratio between over- and underconfidence. Comparable statements exist in algorithmic fairness, where over- and underconfidence were termed \emph{generalized} false positive and negative rates \citep{Pleiss2017,Kleinberg2018}.

\subsection{Calibration Methods}
\label{sec:calibration_methods}

\begin{figure}
\centering
\includegraphics[width=0.48\textwidth]{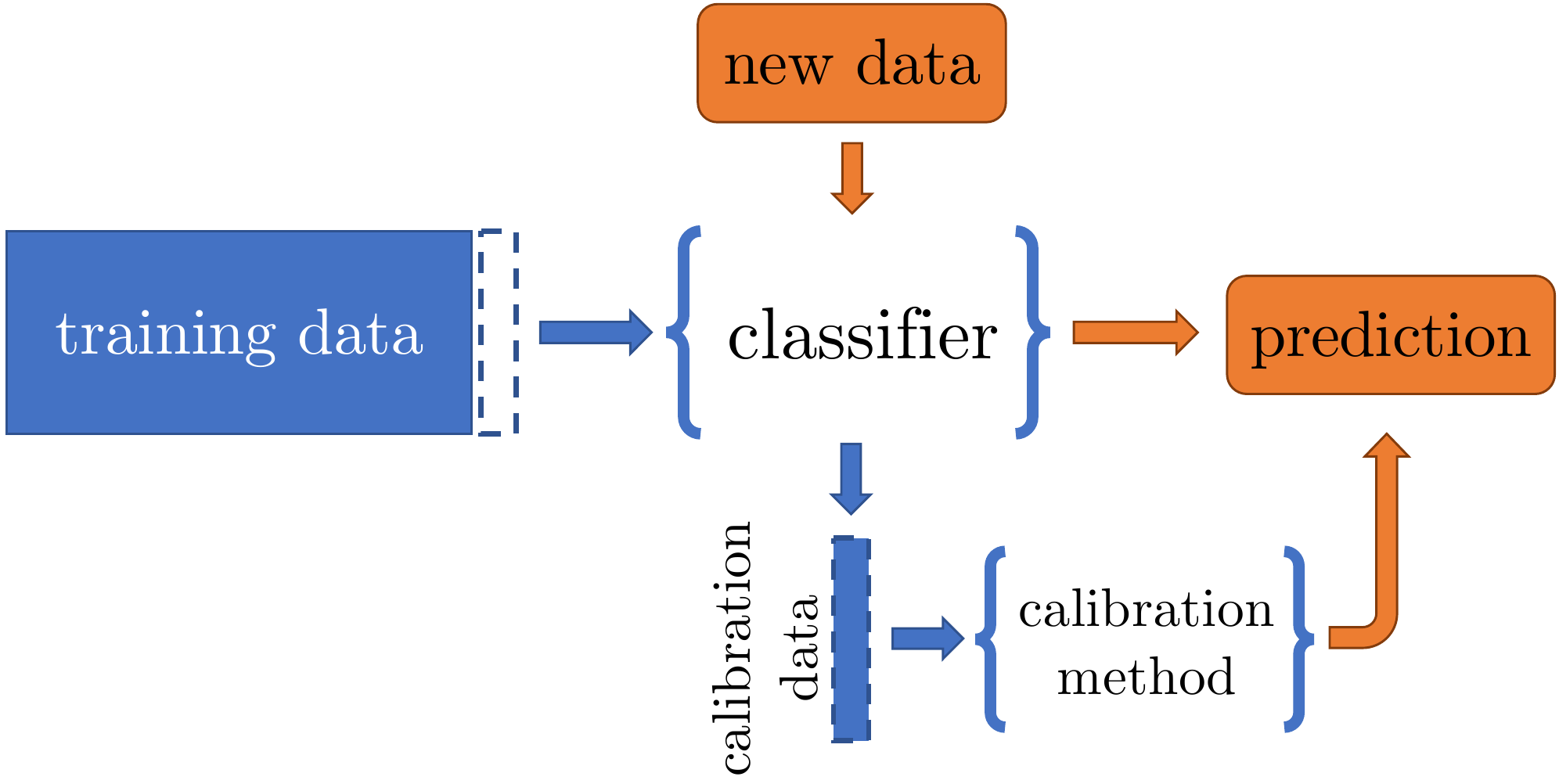}
\caption{\textit{Schematic diagram of calibration.} A fraction of the training data is split off and the remaining data is used for training. The split-off calibration data is classified by the trained model and subsequently used to fit the calibration method (blue). Confidence estimates from the classifier for new data are then adjusted by the calibration method (orange). \label{fig:illustration_calibration}}
\end{figure}

The aim of calibrating a classifier is to transform its output to be closer to the true correctness probability. This is typically done by fitting a calibration method $v$ on a small hold-out set called the \emph{calibration data} (see \Cref{fig:illustration_calibration}). In the following, we describe the most prevalent methods.

\subsubsection{Binary Calibration}

\paragraph{Platt Scaling} \citep{Platt1999,Lin2007} was originally introduced to provide probabilistic output for SVMs.  It is a parametric calibration method, where a logistic regressor is fit to the confidence scores of the positive class such that
\begin{equation*}
v(\rvz)_2 =\left(1+\exp(-a \ervz_2-b)\right)^{-1}
\end{equation*}
for \(a,b \in \mathbb{R}\). This parametric assumption is justified if the scores of each class are normally distributed with identical variance \citep{Kull2017}.

\paragraph{Isotonic Regression} \citep{Zadrozny2002} is a non-parametric approach. 
It assumes a non-decreasing relation between the model confidence $\ervz_2$ of the positive class and its correctness probability $v(\rvz)_2$. A piecewise-constant isotonic function  \(m\) is found by minimizing a squared loss function, resulting in the calibration function \(v(\rvz)_2 = m(\rvz_2) + \rvepsilon\).

\paragraph{Beta Calibration} \citep{Kull2017a} was designed for probabilistic classifiers with output range \(\ervz_2 \in [0,1]\). A family of calibration maps is defined based on the likelihood ratio between two Beta distributions. The calibration map is given by
\begin{equation*}
v(\rvz)_2 = \left(1+\exp(-c)(1-\rvz_2)^b \rvz_2^{-a}\right)^{-1},
\end{equation*}
where \(a,b,c \in \mathbb{R}\) are fit on the calibration data.

\paragraph{Bayesian Binning into Quantiles (BBQ)} \citep{Naeini2015} scores multiple equal-frequency binning models and uses a score weighted average of the accuracy in each bin as a calibration map.  A binning model \(\gM\) is weighted by \(\mathbb{P}(\gM)\mathbb{P}(\mathcal{D} \mid \gM)\), where \(\mathbb{P}(\gM)\) is uniform and the marginal likelihood \(\mathbb{P}(\mathcal{D} \mid \gM)\) can be computed in closed form given parametric assumptions on data generation.

\subsubsection{Multi-class Calibration}

\paragraph{One-vs-All} In order to extend binary calibration methods to multi-class problems, \cite{Zadrozny2002} suggest a one-vs-all approach, training a binary classifier on each split and calibrating subsequently. As most modern classifiers are inherently multi-class, this approach is not feasible anymore. We instead use a one-vs-all approach for the output \(\rvz\) of the multi-class classifier, train a calibration method on each split and average their predictions.

\paragraph{Temperature Scaling} \citep{Guo2017} was introduced as a multi-class extension to Platt scaling for neural networks. For an output logit vector \(\rvz\) of a neural network and a \textit{temperature} parameter \(T >0\), the calibrated confidence is defined as
\begin{equation}
\label{eqn:temperature_scaling}
v(\rvz) =  \sigma \left(\frac{\rvz}{T}\right)=\frac{\exp\left(\frac{\rvz}{T}\right)}{\sum_{j=1}^K \exp\left(\frac{\ervz_j}{T}\right)}.
\end{equation}
The parameter \(T\) is determined by optimizing the NLL. By construction, the accuracy of the classifier is unchanged after scaling. Variants of this method where the factor is replaced by an affine map were shown to be ineffective \citep{Guo2017}.

\section{GAUSSIAN PROCESS CALIBRATION}
\label{sec:gp_calibration}

In the following section, we will outline our non-parametric calibration approach. Our aim is to develop a calibration method, which is inherently \emph{multi-class}, suitable for \textit{arbitrary classifiers}, makes as few assumptions as possible on the shape of the calibration map and can take prior knowledge into account. These desired properties motivate the use of a latent GP.

\begin{figure}
\centering
\includegraphics[width=0.45\textwidth]{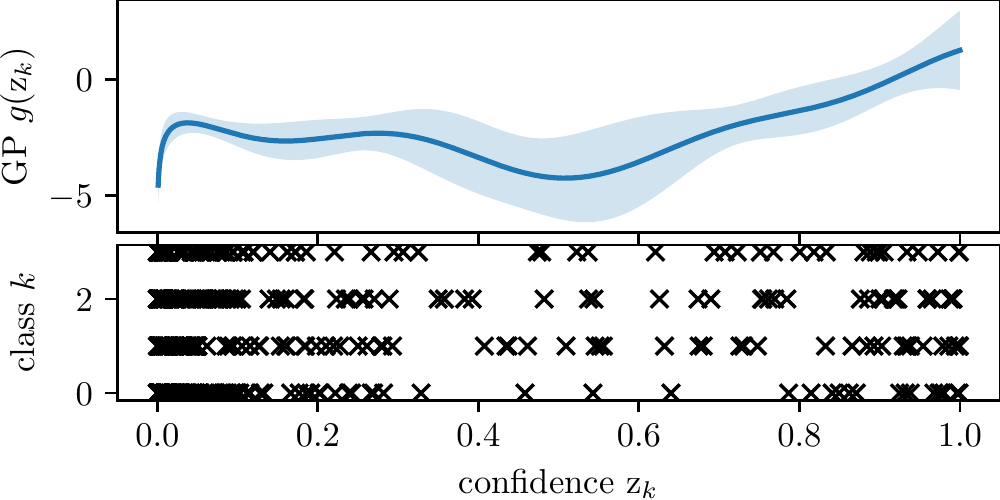}
\caption{\textit{Toy example of multi-class GP calibration.} The top panel shows the resulting latent Gaussian process with prior mean \(\mu(\ervz_k) = \ln(\ervz_k)\) when applying GP calibration to a synthetic calibration data set with four classes and 100 calibration samples. At the bottom, the class confidence scores making up the calibration data are plotted. The \textit{calibration uncertainty} of the latent GP depends on their distribution.\label{fig:multiclass_GP_calib}}
\end{figure}

\paragraph{Definition} Assume a one-dimensional Gaussian process prior over the latent function \(g : \mathbb{R} \rightarrow \mathbb{R}\), i.e. \[g \sim \mathcal{GP}\left(\mu(\cdot), k(\cdot \,, \cdot \mid \bm{\theta})\right)\] with mean function \(\mu\), kernel \(k\) and kernel parameters \(\bm{\theta}\) \citep{Rasmussen2005}. Further, let the calibrated output be given by the \textit{softargmax} inverse link function applied to the latent process evaluated at the model output
\begin{equation*}
\label{eqn:gpcalib_softmax}
v(\rvz)_k = \sigma(g(\ervz_1),\dots, g(\ervz_K))_k = \frac{\exp(g(\ervz_k))}{\sum_{j=1}^K \exp(g(\ervz_j))}.
\end{equation*}
Note the similarity to multi-class Gaussian process classification, which has \(K\) latent functions \citep{Rasmussen2005}. In contrast, we consider \emph{one shared latent function} applied to each component \(\ervz_k\). We use the categorical likelihood
\begin{equation}
\label{eqn:gpcalib_likelihood}
\text{Cat}(\ry \mid v(\rvz))=\prod_{k=1}^{K} \sigma(g(\ervz_1),\dots, g(\ervz_K))_k^{[\ry=k]}
\end{equation}
to obtain a prior on the class prediction. We make the prior assumption that the classifier is already calibrated. This corresponds to either \(\mu(\ervz_k) = \ln(\ervz_k)\) if the inputs are confidence estimates, or to \(\mu(\ervz_k)=\ervz_k\) if the inputs are logits. For specific models other choices may be beneficial, e.g. a linear prior. An example of a latent function for a synthetic data set is shown in \Cref{fig:multiclass_GP_calib}. If the latent function \(g\) is monotonically increasing in its domain, the accuracy of the underlying classifier is unchanged after calibration.

\paragraph{Inference} In order to infer the calibration map, we need to fit the underlying GP based on the confidence scores or logits and labels in the calibration set. Given the likelihood \eqref{eqn:gpcalib_likelihood}, the posterior is not analytically tractable. We use variational inference to approximate the posterior \citep{Girolami2006,Paul2012}. For our method to scale to large data sets we only retain a sparse representation of the inputs, making inference computationally less intensive. We extend an approach by \citet{Hensman2015} to our choice of likelihood. The joint distribution of the data \((\rvz_n, \ry_n)\) and latent variables \(\rvg\) is given by
\begin{align*}
p(\rvy,\rvg) &= p(\rvy \mid \rvg) p(\rvg) = \prod_{n=1}^N p(\ry_n \mid \rvg_n) p(\rvg)\\
&=\prod_{n=1}^N \text{Cat}(\ry_n \mid \sigma(\rvg_n)) \mathcal{N}(\rvg \mid \bm{\mu}, \mSigma_{\rvg}),
\end{align*}
where \(\rvy \in \{1, \dots, K\}^N\), \(\rvg =(\rvg_1, \rvg_2, \dots, \rvg_N)^{\top}\in \R^{NK}\) and \(\rvg_n = (g(\ervz_{n1}), \dots, g(\ervz_{nK}))^{\top} \in \R^K\). The covariance matrix \(\mSigma_{\rvg}\) has block-diagonal structure by independence of the calibration data. If performance is important, a further diagonal assumption can be made. Note that we drop the explicit dependence on \(\rvz_n\) and \(\vtheta\) throughout to lighten the notation. We want to compute the posterior \(p(\rvg \mid \rvy)\). In order to reduce the computational complexity \(\mathcal{O}((NK)^3)\), we define \(M\) inducing inputs \(\rvw \in \R^M\) and inducing variables \(\rvu \in \R^M\). The joint distribution is given by
\begin{equation}
\label{eqn:GP_joint_inducing}
p(\rvg, \rvu) = \mathcal{N}\left(\begin{bmatrix} \rvg\\ \rvu\end{bmatrix} \bigg | \begin{bmatrix} \vmu_{\rvg}\\ \vmu_{\rvu}\end{bmatrix}, 
\begin{bmatrix}
	\mSigma_{\rvg} & \mSigma_{\rvg, \rvu}\\
	\mSigma_{\rvg, \rvu}^{\top} & \mSigma_{\rvu}
\end{bmatrix}
\right).
\end{equation}
The joint distribution factorizes as \(p(\rvy, \rvg, \rvu) = p(\rvy \mid \rvg) p(\rvg \mid \rvu) p(\rvu)\). We aim to find a variational approximation \(q(\rvu)= \mathcal{N}(\rvu \mid \vm, \mS)\) to the posterior \(p(\rvu \mid \rvy)\). For general treatments on variational inference we refer to \citet{Blei2017,Zhang2018}. We find the variational parameters \(\vm\) and \(\mS\), the locations of the inducing inputs \(\rvw\) and the kernel parameters \(\vtheta\) by maximizing a lower bound to the marginal log-likelihood
\begin{equation}
\label{eqn:gpcalib_variational_obj}
\begin{aligned}
\ln p(\rvy) &\geq  \text{ELBO}(q(\rvu))\\
&= \mathbb{E}_{q(\rvu)} \left[\ln p(\rvy \mid \rvu) \right] - \operatorname{KL}\left[ q(\rvu) \| p(\rvu) \right] \\
&\geq \mathbb{E}_{q(\rvu)} \left[ \mathbb{E}_{p(\rvg \mid \rvu)} \left[ \ln p(\rvy \mid \rvg) \right] \right] \\ & \hspace{19.9ex}- \operatorname{KL} \left[ q(\rvu) \| p(\rvu) \right]\\
&=\mathbb{E}_{q(\rvg)} \left[\ln p(\rvy \mid \rvg) \right] - \operatorname{KL}\left[ q(\rvu) \| p(\rvu) \right]\\
&= \sum_{n=1}^N \mathbb{E}_{q(\rvg_n)} \left[\ln p(\ry_n \mid \rvg_n) \right] \\ & \hspace{19.9ex} - \operatorname{KL}\left[ q(\rvu) \| p(\rvu) \right]
\end{aligned}
\end{equation}
where \(q(\rvg) \coloneqq \int p(\rvg \mid \rvu) q(\rvu) \, d\rvu\) is Gaussian and only its \(K\)-dimensional marginals \(q(\rvg_n)=\mathcal{N}(\rvg_n \mid \vphi_n, \mC_n)\) are required to compute the expectation terms. To do so, we use a second order Taylor approximation for \(\ln p(\ry_n \mid \rvg_n)\) and obtain
\begin{multline*}
\mathbb{E}_{q(\rvg_n)} \left[\ln p(\ry_n \mid \rvg_n) \right] \approx \ln p(\ry_n \mid \vphi_n) \\ + \frac{1}{2}  \left(\sigma(\vphi_n)^{\top} \mC_n \sigma(\vphi_n) -\text{diag}(\mC_n)^{\top} \sigma(\vphi_n)\right)
\end{multline*}
which can be computed in \(\mathcal{O}(K^2)\). Computing the KL-divergence term is in \(\mathcal{O}(M^3)\). Therefore, computing the objective \eqref{eqn:gpcalib_variational_obj} has complexity \(\mathcal{O}(NK^2 + M^3)\). Note that this can be remedied through parallelization as all \(N\) expectation terms can be computed independently. The optimization is performed via a gradient-based optimizer and automatic differentiation. We refer to \Cref{sec:supp_gp_calib} of the supplementary material for a more detailed treatment of inference.

\paragraph{Calibration} Given the approximate posterior \(p(\rvg, \rvu \mid \rvy) \approx p(\rvg\mid \rvu) q(\rvu)\), calibrated predictions at new inputs \((\rvz_1, \dots, \rvz_L^*)^\top \in \R^{LN}\) are obtained via
\begin{align*}
p(\rvg_* \mid \rvy) &= \int  p(\rvg_* \mid \rvg, \rvu)p(\rvg, \rvu \mid \rvy) \, d\rvg \, d\rvu\\&\approx \int  p(\rvg_* \mid \rvu)q(\rvu) \,  d\rvu
\end{align*}
which is Gaussian. Mean \(\vmu_{\rvg_*}\) and variance of a latent value \(\rvg_* \in \mathbb{R}^K\) can be computed in \(\mathcal{O}(KM^2)\). The class predictions \(\rvy_*\) are then obtained by marginalization
\begin{equation*}
\label{eqn:GP_cal_predict}
p(\rvy_* \mid \rvy) = \int p(\rvy_* \mid \rvg_*)p(\rvg_* \mid \rvy) \, d\rvg_*
\end{equation*}
via Monte-Carlo integration. While inference and calibration have higher computational cost than in other methods, it is orders of magnitude less than the training cost of the classifier. Furthermore, calibration can be performed in parallel with training in the online setting. We can speed up calibration by approximating the predictive distribution via the GP mean, i.e. \(p(\rvy_* \mid \rvy) \approx p(\rvy_* \mid \vmu_{\rvg_*}) = \sigma(\vmu_{\rvg_*})\).

\section{EXPERIMENTS}
\label{sec:experiments}

We experimentally evaluate our approach against the calibration methods presented in \Cref{sec:calibration_methods}, applied to different classifiers on a range of binary and multi-class computer vision benchmark data sets. Besides CNNs, we are also interested in ensemble methods. All methods and experiments were implemented in Python 3.6. GPcalib was developed based on \texttt{gpflow} \citep{Matthews2017}. Any results reported used a sum kernel consisting of an RBF and a white noise kernel and a diagonal covariance matrix \(\mSigma_{\rvg}\)\ifnum\statePaper=0.\footnote{An implementation of GP calibration and code replicating the experiments is available at \url{<anonymized>}.}
\else
.\footnote{An implementation of GP calibration and code replicating the experiments is available at \url{https://github.com/JonathanWenger/pycalib}.}
\fi

\subsection{Calibration Results}
\label{sec:calibration_results}

We report the average \(\operatorname{ECE}_1\) estimated with 100 bins over 10 Monte-Carlo cross validation runs. We chose a larger number of bins than in previous works, as too few bins typically underestimate the \(\operatorname{ECE}_1\) \citep{Kumar2019}. See \Cref{sec:supp_number_of_bins} of the supplementary material for details. We used the following data sets with indicated train, calibration and test splits:
\begin{itemize}
\setlength\itemsep{0em}
	\item KITTI \citep{Geiger2012,Narr2016}:  Stream-based urban traffic scenes with features \citep{Himmelsbach2009} from segmented 3D point clouds. 8 or 2 classes, train: 16000, calibration: 1000, test: 8000.
	\item PCam \citep{Veeling2018}: Histopathologic scans of (metastatic) tissue from lymph node sections converted to grayscale. 2 classes, train: 22768, calibration: 1000, test: 9000.
	\item MNIST \citep{LeCun1998}: Handwritten digit recognition. 10 classes, train: 60000, calibration: 1000, test: 9000.
	\item CIFAR-100 \citep{Krizhevsky2009}: Image database of tiny color images from the web. 100 classes,, train: 50000, calibration: 1000, test: 9000.
	\item ImageNet 2012 \citep{Russakovsky2015}: Image database of natural objects. 1000 classes, train: 1.2 million, calibration: 1000, test: 10000.
\end{itemize}

\paragraph{Binary Classification} We trained two boosting variants, AdaBoost \citep{Freund1997,Hastie2009} and XGBoost \citep{Chen2016}, two forest variants, Mondrian Forests \citep{Lakshminarayanan2014} and Random Forests \citep{Breiman2001}, and a one layer neural network on the binary KITTI and PCam data sets. We report the average \(\text{ECE}_1\) in \Cref{tab:ece_binary_experiments} in the supplementary material. For binary problems all calibration methods perform similarly with the exception of isotonic regression, which has particularly low calibration error on the KITTI data set. However, due to its piecewise constant calibration map the resulting confidence distribution has a set of singular peaks instead of a smooth distribution. While GPcalib is competitive across data sets and classifiers, it does not outperform isotonic regression. Hence, if exclusively binary problems are of interest a simple calibration method should be preferred. Interestingly, the 1-layer NN trained on KITTI is already well-calibrated, however all calibration methods except isotonic regression and GPcalib increase the \(\text{ECE}_1\).

\paragraph{Multi-class Classification} Besides the aforementioned classification models, which were trained on MNIST, we also calibrated pre-trained CNN architectures\footnote{Pre-trained CNNs were obtained from \url{https://github.com/bearpaw/pytorch-classification} and \url{https://github.com/Cadene/pretrained-models.pytorch}.} on CIFAR-100 and ImageNet. The following CNNs were used: AlexNet \citep{Krizhevsky2012}, VGG19 \citep{Simonyan2014, Liu2015}, ResNet-50, ResNet-152 \citep{He2016}, WideResNet \citep{Zagoruyko2016}, DenseNet-121, DenseNet-BC-190, DenseNet-201 \citep{Huang2017}, InceptionV4 \citep{Szegedy2016}, ResNeXt-29, SE-ResNeXt-50, SE-ResNeXt-101 \citep{Xie2017,Hu2018}, PolyNet \citep{Zhang2017a}, SENet-154 \citep{Hu2018}, PNASNet-5-Large \citep{Liu2018}, NASNet-A-Large \citep{Zoph2018}. All binary calibration methods were extended to the multi-class setting in a one-vs-all manner. Temperature scaling and GPcalib were applied to logits for all CNNs and otherwise directly to probability scores. The average \(\operatorname{ECE}_1\) is shown in \Cref{tab:ece_multiclass_experiments}.
\begin{table*}[ht!]
  \caption{\textit{Multi-class calibration experiments.} Average \(\text{ECE}_1\) of 10 Monte-Carlo cross validation folds on multi-class benchmark data sets. Calibration errors (\(\text{ECE}_1\)) within one standard deviation of lowest per data set and model are printed in bold.}
  \label{tab:ece_multiclass_experiments}
  \centering
{\small    
\begin{filecontents*}{tables/multiclass_ece.csv}
Dataset,Model,uncalECEmean,uncalECEstd,plattECEmean,plattECEstd,isotonicECEmean,isotonicECEstd,betaECEmean,betaECEstd,bbqECEmean,bbqECEstd,tempECEmean,tempECEstd,gpcalibECEmean,gpcalibECEstd
MNIST,AdaBoost,.6121,.0009,.2267,.0137,.1319,.0108,.2222,.0134,.1384,.0104,.1567,.0122,\textbf{.0414},.0085
MNIST,XGBoost,.0740,.0005,.0449,.0021,\textbf{.0176},.0018,\textbf{.0184},.0014,.0207,.0020,.0222,.0015,\textbf{.0180},.0014
MNIST,Mondrian Forest,.2163,.0008,.0357,.0049,.0282,.0021,.0383,.0057,.0762,.0111,\textbf{.0208},.0012,\textbf{.0213},.0020
MNIST,Random Forest,.1178,.0008,.0273,.0039,.0207,.0042,.0259,.0070,.1233,.0005,\textbf{.0121},.0012,.0148,.0021
MNIST,1 layer NN,.0262,.0005,\textbf{.0126},.0031,\textbf{.0140},.0017,.0168,.0018,.0186,.0027,.0195,.0060,.0239,.0023
CIFAR-100,AlexNet,.2751,.0016,.0720,.0090,.1232,.0102,.0784,.0091,.0478,.0058,\textbf{.0365},.0024,\textbf{.0369},.0054
CIFAR-100,WideResNet,.0664,.0009,.0838,.0056,.0661,.0040,.0539,.0031,.0384,.0046,.0444,.0030,\textbf{.0283},.0027
CIFAR-100,ResNeXt-29 (8x64),.0495,.0008,.0882,.0040,.0599,.0062,.0492,.0054,.0392,.0088,.0424,.0015,\textbf{.0251},.0020
CIFAR-100,ResNeXt-29 (16x64),.0527,.0011,.0900,.0041,.0620,.0058,.0520,.0054,.0365,.0078,.0465,.0026,\textbf{.0266},.0018
CIFAR-100,DenseNet-BC-190,.0717,.0006,.0801,.0061,.0665,.0036,.0543,.0036,.0376,.0049,.0377,.0019,\textbf{.0237},.0021
ImageNet,AlexNet,\textbf{.0353},.0025,.1132,.0046,.2937,.0094,.2290,.0078,.1307,.0093,\textbf{.0342},.0035,\textbf{.0357},.0025
ImageNet,VGG19,.0377,.0022,.0965,.0095,.2810,.0131,.2416,.0163,.1617,.0110,\textbf{.0342},.0034,\textbf{.0364},.0018
ImageNet,ResNet-50,.0441,.0025,.0875,.0072,.2724,.0188,.2250,.0122,.1635,.0063,\textbf{.0341},.0018,\textbf{.0335},.0054
ImageNet,ResNet-152,.0545,.0016,.0879,.0066,.2761,.0348,.2201,.0130,.1675,.0064,.0323,.0019,\textbf{.0283},.0038
ImageNet,DenseNet-121,.0380,.0015,.0949,.0061,.2682,.0069,.2297,.0139,.1512,.0099,\textbf{.0329},.0020,.0357,.0040
ImageNet,DenseNet-201,.0410,.0032,.0898,.0111,.2706,.0210,.2189,.0157,.1614,.0116,\textbf{.0324},.0018,.0367,.0058
ImageNet,InceptionV4,.0318,.0025,.0865,.0102,.2900,.0351,.1653,.0138,.1593,.0155,.0462,.0070,\textbf{.0269},.0031
ImageNet,SE-ResNeXt-50,.0440,.0033,.0889,.0101,.2684,.0260,.1789,.0106,.1990,.0176,.0482,.0069,\textbf{.0279},.0055
ImageNet,SE-ResNeXt-101,.0574,.0026,.0853,.0112,.2844,.0218,.1631,.0098,.1496,.0106,.0415,.0039,\textbf{.0250},.0020
ImageNet,PolyNet,.0823,.0031,.0806,.0048,.2590,.0302,.2006,.0124,.1787,.0151,.0369,.0030,\textbf{.0283},.0041
ImageNet,SENet-154,.0612,.0015,.0809,.0106,.3003,.0349,.1582,.0098,.1502,.0127,.0497,.0028,\textbf{.0309},.0030
ImageNet,PNASNet-5-Large,.0702,.0025,.0796,.0081,.3063,.0221,.1430,.0085,.1355,.0101,.0486,.0031,\textbf{.0270},.0018
ImageNet,NASNet-A-Large,.0530,.0027,.0826,.0082,.3265,.0300,.1437,.0085,.1268,.0097,.0516,.0029,\textbf{.0255},.0030
\end{filecontents*}
\csvreader[
	head to column names,
	tabular=llccccccc,
	table head=\toprule &&& \multicolumn{4}{c}{one-vs-all}\\ \cmidrule(r){4-7}Data Set & Model & Uncal. & Platt & Isotonic & Beta & BBQ & Temp. & GPcalib\\ \midrule,
	table foot=\bottomrule,
	separator=comma]
	{tables/multiclass_ece.csv}
	{}
	{\Dataset & \Model & $\uncalECEmean$ & $\plattECEmean$ & $\isotonicECEmean$ & $\betaECEmean$ & $\bbqECEmean$ & $\tempECEmean$ & $\gpcalibECEmean$}}
\end{table*}
While binary methods still perform reasonably well for 10 classes in the case of MNIST and CIFAR-100, they worsen calibration considerably in the case of 1000 classes on ImageNet. Moreover, they also skew the posterior distribution so much that accuracy is heavily affected, disqualifying them from use. Temperature scaling preserves the underlying accuracy of the classifier by definition. Even though GP calibration has no such guarantees, our experiments show little effect on accuracy (see \Cref{tab:accuracy_multiclass_experiments} in the supplementary material). GP calibration performs comparably to other methods on MNIST except for the simple NN and AdaBoost. It does not improve upon calibration for the simple NN, but it is the only method able to handle the large \(\operatorname{ECE}_1\) of AdaBoost. GPcalib calibrates significantly better on CIFAR-100 than all other methods for all CNNs, except AlexNet. On ImageNet GPcalib demonstrates low \(\operatorname{ECE}_1\) within one to two standard deviations of temperature scaling on four CNNs, but outperforms all other calibration methods on the remaining nine evaluated architectures. In particular on higher accuracy CNNs (see \Cref{tab:accuracy_multiclass_experiments}), GPcalib calibrates better. For CNNs which already demonstrate low \(\operatorname{ECE}_1\), such as InceptionV4 and SE-ResNeXt-50, most methods worsen calibration, whereas GPcalib does not. We attribute this desirable behavior, also seen in the binary case, to its prior assumption that the underlying classifier is already calibrated. The increased flexibility of the non-parametric latent map and its prior assumptions allow GPcalib to adapt to various classifiers and data sets.

\begin{figure*}[ht!]
	\centering 
	\includegraphics[width=.9\textwidth]{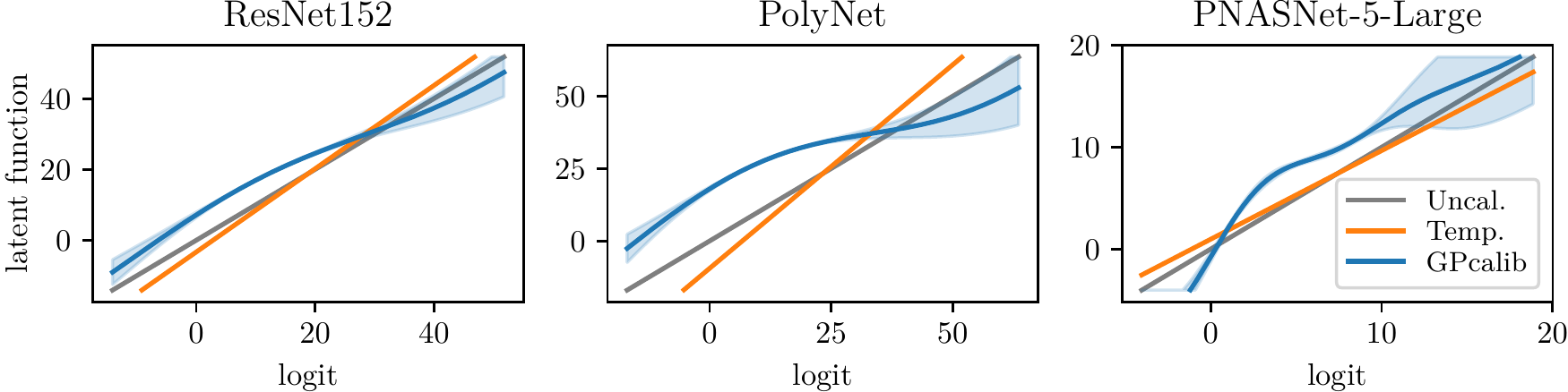}
	\caption{\textit{Non-linear calibration maps in logit-space.} The plot shows latent functions of temperature scaling and GPcalib from a single CV run of our experiments on ImageNet. For PolyNet and PNASNet GPcalib shows a significant decrease in \(\operatorname{ECE}_1\) in \Cref{tab:ece_multiclass_experiments}, corresponding to a higher degree of non-linearity in the latent GP.\label{fig:latent_maps_logit}}
\end{figure*}

\paragraph{Latent Function Visualization}
In order to illustrate the benefit of a non-linear latent function when calibrating, we show some latent functions from our experiments on ImageNet. We compare GPcalib with temperature scaling and no calibration corresponding to the identity map. \Cref{fig:latent_maps_logit} illustrates in logit space how a non-linear latent function allows for lower $\textup{ECE}_1$, when the calibration data necessitates it. When comparing latent functions, note that \(\sigma(g(\rvz) + \textup{const}) = \sigma(g(\rvz))\), i.e. an arbitrary shift of the latent functions on the y-axis corresponds to the same uncertainty estimates. We can also see how the latent GP gives information via its covariance on where the latent function's shape is more certain based on the seen calibration data. For more examples, also for probability scores, we refer to \Cref{sec:supp_latent_functions} of the supplementary material.

\paragraph{Computation Time}
\begin{table}[hb!]
	  \caption{\textit{Computational complexity.} Complexity analysis of evaluation of the optimization objective for parameter inference of the calibration methods and complexity of calibration. \(N\) denotes the size of the calibration data, \(K\) the number of classes, \(M\) the number of inducing points and \(Q\) the number of MC samples. For the mean approximation \(a=1\) with a diagonal covariance and \(a=2\) for the full covariance. Implementation choices are \(M=10\) and \(Q=100\).\label{tab:complexity}}
    \centering
    	{
        \begin{tabular}{lcc} 
        \toprule
        Method & Optim. obj. & Calibration\\ 
        \midrule
        Temp. scal. & \(\mathcal{O}(NK)\) & \(\mathcal{O}(K)\) \\
        GPcalib&\\
        \quad diag. cov.& \(\mathcal{O}(NK+M^3)\) & \(\mathcal{O}(K(M^2 + Q))\) \\
        \quad full cov.& \(\mathcal{O}(NK^2+M^3)\) & \(\mathcal{O}(K^2(M^2 + Q))\) \\
        \quad mean appr.&     \(\mathcal{O}(NK^a+M^3)\) & \(\mathcal{O}(K^aM^2)\) \\
        \bottomrule
        \end{tabular}}
\end{table}
We give a complexity analysis for temperature scaling and GPcalib in \Cref{tab:complexity}.
The cost of evaluation of the optimization objective and calibration for different variants of GP calibration are shown. For a diagonal covariance matrix, evaluating the optimization objective is of similar complexity to temperature scaling, since in general \(NK\) dominates the cubed number of inducing points. However, in our experiments the optimizer converged more slowly for GPcalib than for temperature scaling. We provide wall-clock inference and prediction runtime averaged across models per benchmark data set in \Cref{sec:supp_runtime} of the supplementary material. In practice, for most classifiers the training time is orders of magnitude larger than the time for inferring the latent GP in our calibration method. Calibration is computationally more expensive for GPcalib compared to other methods in part due to the marginalization of the \emph{calibration uncertainty}. This can be reduced by one order of magnitude for data sets with a large number of classes via the mean approximation with practically no effect on the \(\text{ECE}_1\). The resulting time taken for calibration is about one-order of magnitude more than temperature scaling. In our experiments this was at least two orders of magnitude less than the prediction time of the classifier.

\section{CONCLUSION}
\label{sec:conclusion}
In this paper we proposed a novel multi-class calibration method for arbitrary classifiers based on a latent Gaussian process, inferred via variational inference. We evaluated different calibration methods for a range of classifiers often employed in computer vision and robotics on a collection of benchmark data sets. Our method demonstrated strong performance across different model and data set combinations and performed particularly well for large-scale neural networks. In \Cref{thm:calibration_overconfidence} we linked calibration to concepts from active learning. We conclude with a motivating example for the possible impact of calibration on active learning and outline future research directions.

\paragraph{Active Learning} 
\begin{figure}[t!]
\centering
\includegraphics[width=.47\textwidth]{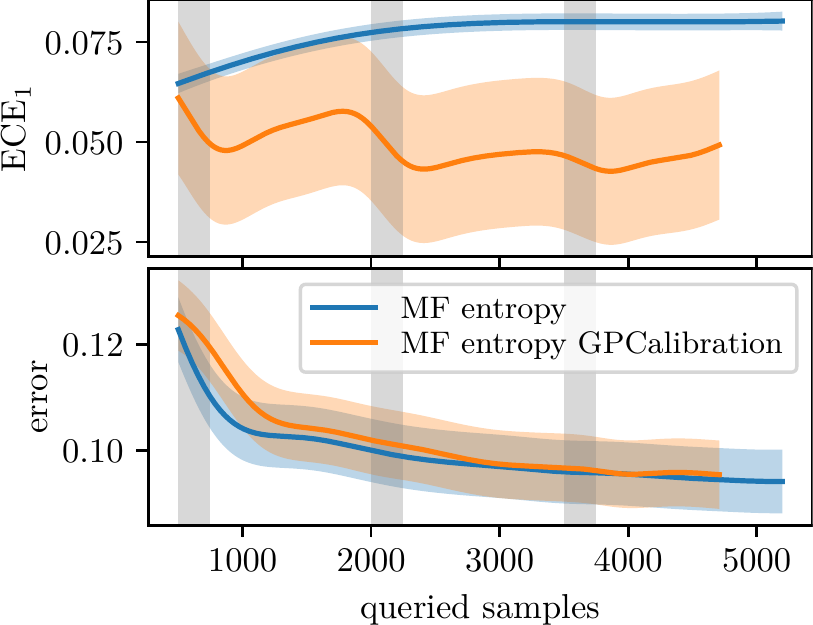}
\caption{\textit{Active learning and calibration.} \(\text{ECE}_1\) and classification error for two Mondrian forests trained online on labels obtained via an entropy query strategy on the KITTI data set. One forest is calibrated in regularly spaced intervals with GPcalib (gray). A GP regression up to the average number of queried samples across folds is shown. The calibrated forest queries \(\sim 10\%\) less labels, while reaching comparable accuracy.\label{fig:al_ece_error}}
\end{figure}

We hypothesize that calibration could improve active learning when querying based on uncertainty. We trained two Mondrian forests on the multi-class KITTI data set. These are well-suited for the online setting as they have the same distribution whether trained online or in batch. We randomly shuffled the data 10 times and requested samples based on an entropy query strategy with a threshold of 0.25. Any samples above the threshold are used for training or calibration. Both forests are trained for 500 samples and subsequently one uses 250 samples exclusively for calibration in regularly spaced intervals. We report the \(\operatorname{ECE}_1\) and classification error in \Cref{fig:al_ece_error}. The calibration initially incurs a penalty on accuracy, as fewer samples are used for training. This is remedied over time through more efficient querying. The same accuracy is reached after a pass through the entire data set while \emph{querying less samples overall}. This can be explained by calibration adjusting over- and underconfidence to the ratio determined by \Cref{thm:calibration_overconfidence}. Here, underconfidence is reduced, leading to less querying of uninformative samples. This is a promising result, but further research questions arise, regarding the size of the pre-training batch, the condition when calibration should be done, and the number of samples used for calibration. We believe that there is a trade-off, similar to an explore-exploit strategy, between classifier training and uncertainty calibration.

\paragraph{Future Directions}
Our proposed calibration approach is worth extending in the following directions: (a) forcing a monotone latent Gaussian process \citep{Riihimaeki2010, Agrell2019} provably preserves the accuracy of the underlying classifier; (b) extending our method to the online setting \citep{Bui2017} allows for continuous calibration; (c) using our method for what we call \emph{active calibration}, the concept of using an active learning query strategy, which switches between requesting samples for model training and uncertainty calibration based on the uncertainty of the latent Gaussian process.

\ifnum\statePaper=1
\subsubsection*{Acknowledgements}
JW gratefully acknowledges financial support by the European Research Council through ERC StG Action 757275 / PANAMA; the DFG Cluster of Excellence ``Machine Learning - New Perspectives for Science'', EXC 2064/1, project number 390727645; the German Federal Ministry of Education and Research (BMBF) through the Tübingen AI Center (FKZ: 01IS18039A); and funds from the Ministry of Science, Research and Arts of the State of Baden-Württemberg. JW is grateful to the International Max Planck Research School for Intelligent Systems (IMPRS-IS) for support. RT gratefully acknowledges the Helmholtz Artificial Intelligence Cooperation Unit (HAICU), which partly supported this work. 

The authors thank the anonymous reviewers for helpful comments on an earlier version of this manuscript. In particular we appreciate the suggestions on inference using the mean estimate of the latent Gaussian process and regarding the definition of over- and underconfidence.
\fi

\bibliographystyle{unsrtnat}
\bibliography{nonparametric_calibration.bib}

\clearpage

\renewcommand{\thesection}{S\arabic{section}}
\renewcommand{\thesubsection}{\thesection.\arabic{subsection}}
\renewcommand{\theHsection}{S\arabic{section}} 

\newcommand{\beginsupplementary}{%
	\setcounter{table}{0}
	\renewcommand{\thetable}{S\arabic{table}}%
	\setcounter{figure}{0}
	\renewcommand{\thefigure}{S\arabic{figure}}%
	\setcounter{section}{0}
}
\beginsupplementary

\thispagestyle{empty}
\twocolumn[
\aistatstitle{Supplementary Material:\\ \papertitle{}}
]

This is the supplementary material for the paper: ``\papertitle{}''\ifnum\statePaper=0, by Author 1, Author 2 and Author 3.
\else, by \firstauthor{}, \secondauthor{} and \thirdauthor{}. 
\fi

\section{OVER- AND UNDERCONFIDENCE}
\label{sec:supp_over_underconfidence}
The notions over- and underconfidence
\begin{equation*}
o(f) = \mathbb{E}\left[\hat{\rz} \mid \hat{\ry} \neq \ry \right] \quad u(f) = \mathbb{E}\left[1-\hat{\rz} \mid \hat{\ry} = \ry\right],
\end{equation*}
as introduced in \eqref{eqn:over_underconfidence}, quantify the amount of information contained in the uncertainty estimate of a classifier \(f\) about the true class. Note, that their definitions are decoupled from the accuracy of \(f\).

\subsection{Definition Subtleties}
Differing from the corresponding intuitive notions, a classifier can simultaneously be over- and underconfident to varying degree. In particular \(o(f) > 0\) does not imply \(u(f) = 0\), and neither does the reverse. Consider the following, where the true posterior \(p(\ry \mid \rvx)\) assigns \(75\%\) confidence to one class and the remaining \(25\%\) uniformly across all other classes. Assume now the classifier assigns \(90\%\) confidence to the given class and \(10\%\) uniformly to the remaining classes on the entire input space. Intuition would dictate this classifier to not necessarily be underconfident, but by definition \(u(f) = 0.1\) and \(o(f) = 0.9\). This is because over- and underconfidence are not a statistical distance between posterior distributions, but describe the difference between the classifier's posterior distribution and an unobserved deterministic underlying relationship \((\rvx, \ry)\). Hence, aleatoric and epistemic uncertainty contained in the data distribution influence the over- and underconfidence of a classifier. This can be seen by computing the over- and underconfidence for the true posterior distribution in the above example giving \(u(f) = 0.25\) and \(o(f) = 0.75\). Obtaining lower over- or underconfidence in this case is only possible by sacrificing one or the other.

\subsection{Proof of Theorem \ref{thm:calibration_overconfidence}}
\label{sec:supp_proofs}

We give a proof for the calibration error bound to the weighted absolute difference between over- and underconfidence as stated in \Cref{thm:calibration_overconfidence} below.

\begin{proof}
By linearity of expectation and the law of total expectation it holds that
\begin{align*}
\E{\hat{\rz}} &= \E{\hat{\rz} + \E{1_{\hat{\ry} = \ry} \mid \hat{\rz}} - \E{1_{\hat{\ry} = \ry} \mid \hat{\rz}}}\\
&= \E{\hat{\rz} - \E{1_{\hat{\ry} = \ry} \mid \hat{\rz}}} + \mathbb{P}(\hat{\ry} = \ry).\\
\intertext{Conversely, by decomposing the average confidence we have}
\E{\hat{\rz}} &= \E{\hat{\rz} \mid \hat{\ry} \neq \ry} \mathbb{P}(\hat{\ry} \neq \ry) + \E{\hat{\rz} \mid \hat{\ry} = \ry}\mathbb{P}(\hat{\ry} = \ry)\\
&= \E{\hat{\rz} \mid \hat{\ry} \neq \ry} \mathbb{P}(\hat{\ry} \neq \ry) + \\
&(1-\E{1-\hat{\rz} \mid \hat{\ry} = \ry})\mathbb{P}(\hat{\ry} = \ry)\\
&= o(f)\mathbb{P}(\hat{\ry} \neq \ry) + (1-u(f))\mathbb{P}(\hat{\ry} = \ry).
\end{align*}
Combining the above we obtain
\begin{equation*}
\E{\hat{\rz} - \E{1_{\hat{\ry} = \ry} \mid \hat{\rz}}} = o(f)\mathbb{P}(\hat{\ry} \neq \ry) -u(f)\mathbb{P}(\hat{\ry} = \ry).
\end{equation*}
Now, since \(h(x) = \abs{x}^p\) is convex for \(1 \leq p < \infty\), we have by Jensen's inequality
\begin{equation*}
\abs{\E{\hat{\rz} - \E{1_{\hat{\ry} = \ry} \mid \hat{\rz}}}}^p \leq \E{\abs{\hat{\rz} - \E{1_{\hat{\ry} = \ry} \mid \hat{\rz}}}^p}
\end{equation*}
and finally by H\" older's inequality with \(1 \leq p < q \leq \infty\) it follows that
\begin{align*}
\textup{ECE}_p &= \E{\abs{\hat{\rz} - \E{1_{\hat{\ry} = \ry} \mid \hat{\rz}}}^p}^{\frac{1}{p}}\\
 &\leq \E{\abs{\hat{\rz} - \E{1_{\hat{\ry} = \ry} \mid \hat{\rz}}}^q}^{\frac{1}{q}}= \textup{ECE}_q,
\end{align*}
which concludes the proof.
\end{proof}

\section{DETAILED INFERENCE AND CALIBRATION}
\label{sec:supp_gp_calib}

We give a more detailed exposition of GP calibration inference. We begin by describing the derivation of the bound on the marginal log-likelihood in \eqref{eqn:gpcalib_variational_obj}.

\subsection{Bound on the Marginal Log-Likelihood}

This subsection follows \cite{Hensman2015} and is adapted for our specific inverse link function and likelihood. Consider the following bound, derived by marginalization and Jensen's inequality.
\begin{equation}
\begin{aligned}
\label{eqn:GP_inducing_conditional_bound}
\ln p(\rvy \mid \rvu) &= \ln \mathbb{E}_{p(\rvg \mid \rvu)} \left[ p(\rvy \mid \rvg) \right]\\
&\geq \mathbb{E}_{p(\rvg \mid \rvu)} \left[ \ln p(\rvy \mid \rvg) \right]
\end{aligned}
\end{equation}
We then substitute \eqref{eqn:GP_inducing_conditional_bound} into the lower bound to the evidence (ELBO) as follows
\begin{equation}
\label{eqn:supp_gpcalib_variational_obj}
\begin{aligned}
\ln p(\rvy) &= \operatorname{KL} \left[ q(\rvu) \| p(\rvu \mid \rvy) \right] + \text{ELBO}(q(\rvu))\\
&\geq  \text{ELBO}(q(\rvu))\\
&= \mathbb{E}_{q(\rvu)} \left[\ln p(\rvy, \rvu) \right] - \mathbb{E}_{q(\rvu)} \left[\ln q(\rvu) \right]\\
&= \mathbb{E}_{q(\rvu)} \left[\ln p(\rvy \mid \rvu) \right] - \operatorname{KL} \left[ q(\rvu) \| p(\rvu) \right] \\
&\geq \mathbb{E}_{q(\rvu)} \left[ \mathbb{E}_{p(\rvg \mid \rvu)} \left[ \ln p(\rvy \mid \rvg) \right] \right] \\ & \hspace{19.9ex}- \operatorname{KL} \left[ q(\rvu) \| p(\rvu) \right]\\
&=\mathbb{E}_{q(\rvg)} \left[\ln p(\rvy \mid \rvg) \right] - \operatorname{KL} \left[ q(\rvu) \| p(\rvu) \right]\\
&= \sum_{n=1}^N \mathbb{E}_{q(\rvg_n)} \left[\ln p(\ry_n \mid \rvg_n) \right]\\ & \hspace{19.9ex} - \operatorname{KL} \left[ q(\rvu) \| p(\rvu) \right],
\end{aligned}
\end{equation}
where \(q(\rvg) \coloneqq \int p(\rvg \mid \rvu) q(\rvu) \, d\rvu\) and the last equality holds by independence of the calibration data. By \eqref{eqn:GP_joint_inducing} and the properties of Gaussians we obtain
\begin{equation*}
p(\rvg \mid \rvu) = \mathcal{N}(\rvg \mid \vmu_{\rvg \mid \rvu}, \mSigma_{\rvg \mid \rvu})
\end{equation*} such that
\begin{align*}
\vmu_{\rvg \mid \rvu} &= \vmu_{\rvg} + \mSigma_{\rvg, \rvu}\mSigma_{\rvu}^{-1}(\rvu - \bm{\mu}_{\rvu})\\
\mSigma_{\rvg \mid \rvu} &= \mSigma_{\rvg}- \mSigma_{\rvg, \rvu}\mSigma_{\rvu}^{-1}\mSigma_{\rvg, \rvu}^{\top}.
\end{align*}
Let \(q(\rvu) = \mathcal{N}(\rvu \mid \vm, \mS)\) and \(\mA \coloneqq \mSigma_{\rvg, \rvu}\mSigma_{\rvu}^{-1}\), then
\begin{align*}
q(\rvg) &\coloneqq \int \underbrace{p(\rvg \mid \rvu) q(\rvu)}_{q(\rvg, \rvu)} \, d\rvu\\
&= \mathcal{N}(\rvg \mid \vmu_{\rvg} + \mA(\vm - \vmu_u), \, \mSigma_{\rvg} + \mA (\mS-\mSigma_{\rvu}) \mA^{\top}).
\end{align*}
as \(q(\rvg, \rvu)\) is normally distributed. To compute the expectations in \eqref{eqn:supp_gpcalib_variational_obj} we only need to consider the \(K\)-dimensional marginals 
\begin{equation*}
q(\rvg_n)= \int p(\rvg_n \mid \rvu) q(\rvu) \, d\rvu = \mathcal{N}(\rvg_n \mid \bm{\varphi}_n, \mC_n).
\end{equation*}

\subsection{Approximation of the Expectation Terms}
In order to obtain the variational objective \eqref{eqn:supp_gpcalib_variational_obj} we need to compute the expected value terms for our intractable likelihood \eqref{eqn:gpcalib_likelihood}. To do so, we use a second order Taylor approximation of
\begin{equation*}
h(\rvg_n) \coloneqq \ln p(\ry_n \mid \rvg_n) = \ln \frac{\exp(\ervg_{n\ry_n})}{\sum_{k=1}^K \exp(\ervg_{nk})}
\end{equation*}
at \(\rvg_n=\bm{\varphi}_n\). The Hessian of the log-softargmax is given by
\begin{align*}
D_{\rvg_n}^2 h(\rvg_n) &= D_{\rvg_n}^2  \ln \sigma(\rvg_n)_{\ry_n}\\
 &=\sigma(\rvg_n) \sigma(\rvg_n)^{\top} - \text{diag}(\sigma(\rvg_n)).
\end{align*}
Note this expression does not depend on \(\ry_n\). We obtain by using \(\vx^{\top}\mM \vx = \tr(\vx^{\top}\mM\vx)\), the linearity of the trace and its invariance under cyclic permutations, that
\begin{equation*}
\begin{split}
\mathbb{E}&_{q(\rvg_n)} \left[\ln p(\ry_n \mid \rvg_n) \right] = \mathbb{E}_{q(\rvg_n)} \left[h(\rvg_n) \right]\\
&\approx \mathbb{E}_{q(\rvg_n)} \Big[ h(\bm{\varphi}_n) + D_{\rvg_n} h(\bm{\varphi}_n)^{\top}(\rvg_n - \bm{\varphi}_n) \\
&\hspace{16.5ex} + \frac{1}{2} (\rvg_n-\bm{\varphi}_n)^{\top} D_{\rvg_n}^2 h(\bm{\varphi}_n) (\rvg_n-\bm{\varphi}_n)\Big]\\
&= h(\bm{\varphi}_n) +  \frac{1}{2} \mathbb{E}_{q(\rvg_n)} \Big[ (\rvg_n - \bm{\varphi}_n)^{\top}\big(\sigma(\bm{\varphi}_n)\sigma(\bm{\varphi}_n)^{\top}\\
&\hspace{24ex} - \text{diag}(\sigma(\bm{\varphi}_n))\big)(\rvg_n - \bm{\varphi}_n) \Big]\\
&= h(\bm{\varphi}_n) +  \frac{1}{2} \tr \Big[\mathbb{E}_{q(\rvg_n)} \big[(\rvg_n - \bm{\varphi}_n) (\rvg_n - \bm{\varphi}_n)^{\top}\Big]\\
& \hspace{20ex}\big(\sigma(\bm{\varphi}_n)\sigma(\bm{\varphi}_n)^{\top} - \operatorname{diag}(\sigma(\bm{\varphi}_n))\big)]\\
&= h(\bm{\varphi}_n)+  \frac{1}{2} \tr[\mC_n \left(\sigma(\bm{\varphi}_n)\sigma(\bm{\varphi}_n)^{\top} - \operatorname{diag}(\sigma(\bm{\varphi}_n))\right)]\\
&= h(\bm{\varphi}_n) + \frac{1}{2} \big(\tr[\sigma(\bm{\varphi}_n)^{\top} \mC_n \sigma(\bm{\varphi}_n)]\\
&\hspace{24ex}- \tr[\mC_n \text{diag}(\sigma(\bm{\varphi}_n))]\big)\\
&= h(\bm{\varphi}_n) + \frac{1}{2}  \big(\sigma(\bm{\varphi}_n)^{\top} \mC_n \sigma(\bm{\varphi}_n)-\text{diag}(\mC_n)^{\top} \sigma(\bm{\varphi}_n)\big),
\end{split}
\end{equation*}
which can be computed in \(\mathcal{O}(K^2)\). This is apparent when expressing the term inside the parentheses as a double sum over \(K\) terms.

\section{EXPERIMENT DETAILS}
\label{sec:supp_experiments}

In this section, we elaborate on the experiments in \Cref{sec:experiments}. We discuss the approximation to the expected calibration error, hyperparameter choices and give runtime measurements and classification accuracy of the performed calibration experiments. Finally, we show some additional visualizations of latent functions and reliability diagrams.

\subsection{Choice of Number of Bins}
\label{sec:supp_number_of_bins}
In practice, we estimate the calibration error as suggested by \cite{Naeini2015} by introducing a fixed uniform binning  \(0=\vartheta_0 < \vartheta_1 < \dots < \vartheta_B=1\) such that
\begin{equation}
\label{eqn:ece_approximation}
\textup{ECE}_p \approx \frac{1}{B} \left(\sum_{b=1}^B \abs{\bar{\hat{\rz}}_b - \textup{acc}_b}^p\right)^{\frac{1}{p}},
\end{equation}
where \[\bar{\hat{\rz}}_b = \frac{1}{N_b}\sum_{\vartheta_{b-1} < \hat{\rz} \leq \vartheta_{b}} \hat{\rz}\] is the mean confidence in bin \(b\), \[\textup{acc}_b= \frac{1}{N_b}\sum_{\vartheta_{b-1} < \hat{\rz} \leq \vartheta_{b}} 1_{\hat{\ry}=\ry}\] the accuracy in bin \(b\) and \(N_b\) the number of samples in bin \(b\), such that \(N=\sum_{b=1}^B N_b\). In previous work introducing the calibration error \citep{Naeini2015} and in \citep{Guo2017} the discretization of the expected calibration error \(\operatorname{ECE}_1\) uses 15 equally spaced bins.

As described in the supplementary material in \citep{Guo2017}, the empirical estimate is a good approximation to the expected calibration error for \(N\) and \(B\) sufficiently large. However, in the infinite sample case the larger the bin size \(B\) the tighter the empirical estimate lower bounds the \(\operatorname{ECE}_1\) \citep{Kumar2019}. This is due to the fact that over- and underestimation of uncertainty within one bin cancel each other out. Hence, \emph{using too few bins can underestimate the expected calibration error}. \cite{Kumar2019} also observe this in practice. This phenomenon is particularly prevalent for CNNs as most of their confidence predictions fall into one or two bins for \(B=15\). This is also the case in our experiments and can be seen in the histograms in \Cref{fig:reldiagram_resnet152,fig:reldiagram_polynet,fig:reldiagram_pnasnet5large}. We observed the aforementioned underestimation of calibration error with 15 bins for some data set and model combinations in our experiments. To mitigate this problem we deliberately chose \(B=100\) in this work. The number of bins is limited by the number of test samples we have available, as within-bin-variance increases with the number of bins. Further work needs to be done to determine the properties of the estimator to \(\operatorname{ECE}_p\) and the ideal discretization for a given number of samples.

\subsection{Implementation and Hyperparameters}
All experiments were performed using the \texttt{pycalib} package available at
\begin{center} \url{https://github.com/JonathanWenger/pycalib}. \end{center}
When calibrating models returning logits, we used a sum kernel for the one-dimensional latent Gaussian process given by
\begin{align*}
k(x, x') &= k_{\text{RBF}}(x, x') + k_{\text{noise}}(x, x')\\
&= \sigma^2 \operatorname{exp}\left(- \frac{\norm{x - x'}^2}{2l^2}\right) + \sigma_{\text{noise}}^2 \delta_{x, x'}
\end{align*}
where \(k_\text{RBF}\) is an exponentiated quadratic and \(k_{\text{noise}}\) a white noise kernel. The kernel parameters were initialised as \(\sigma=1, l=10\) and \(\sigma_{\text{noise}}=0.01\). For GP inference we used \(M=10\) inducing points. The \texttt{gpflow} package with the \texttt{scipy} implementation of the L-BFGS optimizer was used to find inducing points and kernel hyperparameters in the variational inference procedure. For calibration we used \(Q=100\) Monte-Carlo samples to compute the posterior distribution.

\subsection{Binary Experiments}
\label{sec:supp_binary_experiments}

For the binary calibration experiments we used the following data sets with indicated train, calibration and test splits:
\begin{itemize}
\setlength\itemsep{0em}
	\item KITTI \citep{Geiger2012,Narr2016}:  Stream-based urban traffic scenes with features \citep{Himmelsbach2009} from segmented 3D point clouds. 8 or 2 classes, train: 16000, calibration: 1000, test: 8000.
	\item PCam \citep{Veeling2018}: Histopathologic scans of (metastatic) tissue from lymph node sections converted to grayscale. 2 classes, train: 22768, calibration: 1000, test: 9000.
\end{itemize}
The resulting average calibration error across random samples of calibration data sets is shown in \Cref{tab:ece_binary_experiments}. Isotonic regression performed the best in terms of calibration on KITTI. However most methods performed well and often within one standard deviation of each other for both binary data sets. Hence, for binary problems a simple binary calibration method may suffice.

\subsection{Multi-class Experiments}
We provide more detailed calibration results of our multi-class experiment explained in \Cref{sec:calibration_results} in \Cref{tab:ece_std_multiclass_experiments}. We show the average calibration error (\(\text{ECE}_1\)), including standard deviation, of the presented calibration methods and the GPcalib mean approximation on all data set and model combinations.

\subsection{Accuracy}
The accuracy from the binary and multi-class experiments described in \Cref{sec:experiments} is given in \Cref{tab:accuracy_binary_experiments} and \Cref{tab:accuracy_multiclass_experiments}, respectively. For the binary experiments accuracy is mostly unaffected across classifiers and even improves in some instances. Only Bayesian binning into quantiles suffers from a noticable drop in accuracy for random forests. Somewhat surprisingly, for the simple neural network all binary methods actually improve upon accuracy.

In the multi-class case we see that accuracy is severely affected for binary methods extended in a one-vs-all fashion for the ImageNet data set, disqualifying them from use. Both temperature scaling and GP calibration preserve accuracy across models and data sets.

\subsection{Wall-Clock Runtime}
\label{sec:supp_runtime}

We provide wall-clock runtime for each data set considered in our experiments from \Cref{sec:experiments}. As the runtime was very similar across classifiers we show average runtime per data set across models in \Cref{tab:wall_clock_runtime}. Note that wall-clock runtime is highly dependent on the specific machine used for computation. In our case we used a 12-core desktop computer with a GeForce RTX 2080 Ti graphics card for all experiments in this work. Time for inference and calibration scales close to linear with the number of classes for almost all calibration methods. There seems to be some unavoidable overhead for calibration irrespective of classes as can be seen when looking at binary problems. GPcalib takes more time for parameter inference than other methods. This is due to the fact that we need to perform approximate inference and since GPcalib is the only method taking calibration uncertainty into account. Potentially speed-up via the use of GPUs for accelerated matrix computations for our \texttt{gpflow}-based implementation of GPcalib is possible. However, in practice the times taken for inference and calibration are significantly less than those for training and prediction of the underlying classifier. Hence, \emph{a classifier can be calibrated with little added time cost}.

\begin{table*}
  \caption{\textit{Average wall-clock runtime of experiments.} We show average time in seconds of inference on the calibration sets and calibration on the test sets. Times are averaged for each data set across classifiers and 10 Monte-Carlo cross validation folds, since variance between classifiers on the same data set was small.}
  \label{tab:wall_clock_runtime}
  \centering
{\small
\begin{filecontents*}{tables/experiments_fit_calib_time.csv}
Mode,Dataset,platttimemean,isotonictimemean,betatimemean,bbqtimemean,temptimemean,gpcalibtimemean,gpcalibapproxtimemean
Inference,KITTI,0.0017,0.0004,0.0032,0.0994,0.0053,1.0517,1.0441
,PCam,0.0020,0.0007,0.0026,0.0923,0.0063,1.0653,1.0492
,MNIST,0.0245,0.0055,0.0478,0.8589,0.0129,2.2112,2.1596
,CIFAR-100,0.3181,0.0602,0.2052,9.0458,0.0277,20.6564,20.4714
,ImageNet,3.0911,0.9436,2.2291,61.3060,0.3395,259.7305,258.0205
,,,,,,,,
Calibration,KITTI,0.0002,0.0003,0.0022,0.1334,0.0006,0.1625,0.1227
,PCam,0.0003,0.0004,0.0021,0.1502,0.0008,0.1610,0.1171
,MNIST,0.0043,0.0046,0.1126,1.5180,0.0026,0.2610,0.1483
,CIFAR-100,0.1276,0.1269,0.3202,14.8716,0.0210,2.4482,0.3367
,ImageNet,8.3651,8.3005,12.7822,113.6108,0.2369,51.3806,3.1317
\end{filecontents*}
\csvreader[
	head to column names,
	tabular=llccccccc,
	table head=\toprule && \multicolumn{4}{c}{one-vs-all}\\ \cmidrule(r){3-6}Mode & Data Set & Platt & Isotonic & Beta & BBQ & Temp. & GPcalib & GPcalib \\ &&&&&&&& mean appr.\\ \midrule,
	table foot=\bottomrule,
	separator=comma]
	{tables/experiments_fit_calib_time.csv}
	{}
	{\Mode & \Dataset & $\platttimemean$ & $\isotonictimemean$ & $\betatimemean$ & $\bbqtimemean$ & $\temptimemean$ & $\gpcalibtimemean$ & $\gpcalibapproxtimemean$}}
\end{table*}

\subsection{Additional Latent Function Plots}
\label{sec:supp_latent_functions}
\begin{figure*}
	\centering 
	\begin{subfigure}[b]{0.9\textwidth}
		\includegraphics[width=\textwidth]{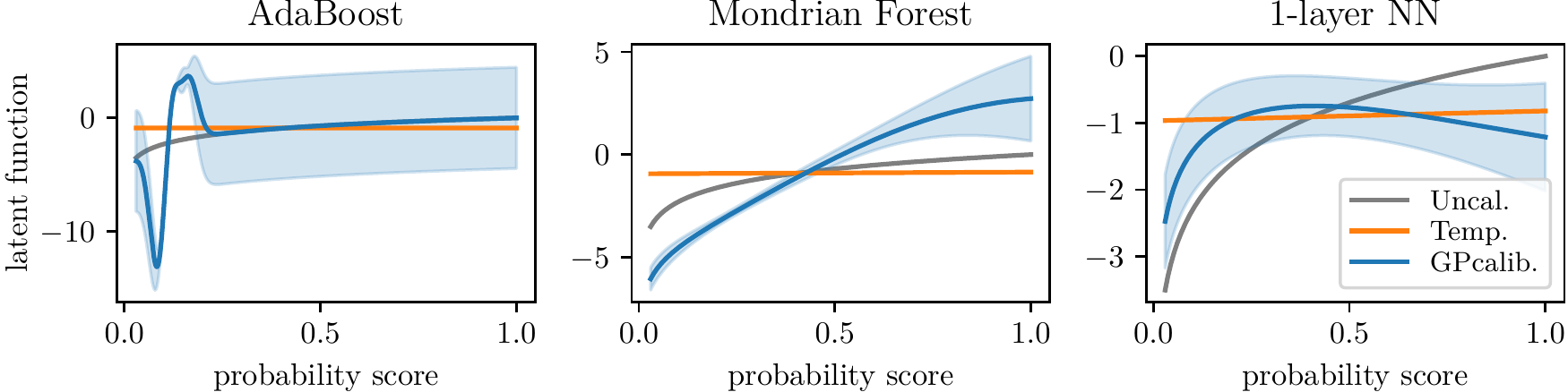}
	\end{subfigure}\\
	\vspace{1.5em}
	\begin{subfigure}[b]{0.6\textwidth}
		\includegraphics[width=\textwidth]{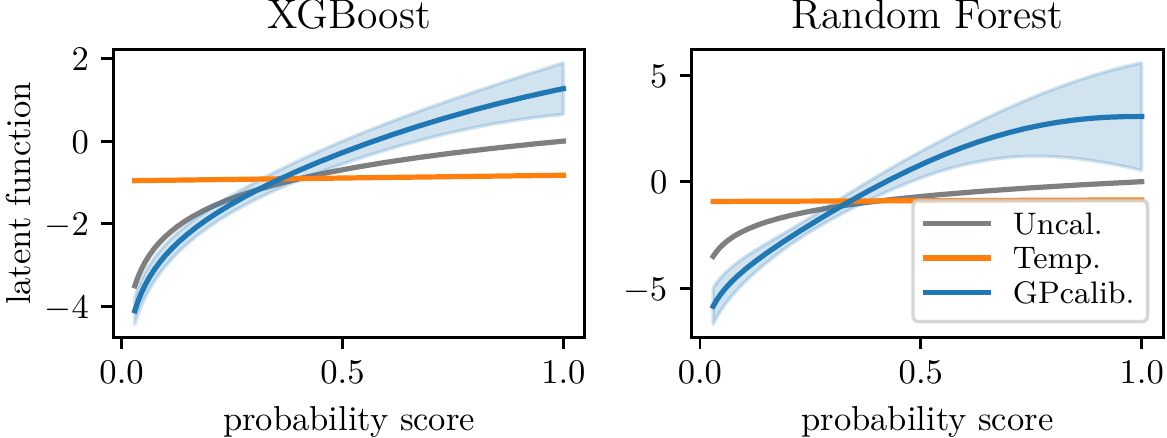}
	\end{subfigure}
	\caption{\textit{Illustration of latent functions on probability scores.} Latent functions of GPcalib and temperature scaling for probability scores from classifiers on MNIST. GPcalib uses the \(\ln\) prior corresponding to the prior assumption of the classifier being calibrated. For AdaBoost we can see in \Cref{tab:ece_multiclass_experiments}, that remedying the large calibration error is only handled well by GPcalib. This corresponds to a large deviation from the prior in latent space.\label{fig:latent_maps_probs}}
\end{figure*}

Further examples of latent calibration maps from GPcalib and temperature scaling are given in this section. \Cref{fig:latent_maps_probs} shows latent functions from a single CV run of the MNIST experiments in \Cref{sec:calibration_results}.When applying GPcalib to classifiers outputting probability scores, the \(\ln\) prior corresponds to the assumption of the underlying model being calibrated. Note, that temperature scaling was not designed to be used on probability scores, but on logits, this causes its latent function to have very small slope. Both in the case of probability scores and logits, by definition of GPcalib and temperature scaling any constant shift of the latent function results in the same uncertainty estimates. The shown plots should be interpreted with this in mind. For models which are very miscalibrated, the latent function of GPcalib demonstrates a high degree of non-linearity and deviation from the \(\ln\) prior, e.g. in the case of AdaBoost. When the underlying model is already close to being calibrated as in the case of the 1-layer neural network, the resulting latent GP does not deviate very much from the prior mean. 

\begin{figure*}
	\centering
	\begin{subfigure}[b]{0.9\textwidth}
		\includegraphics[width=\textwidth]{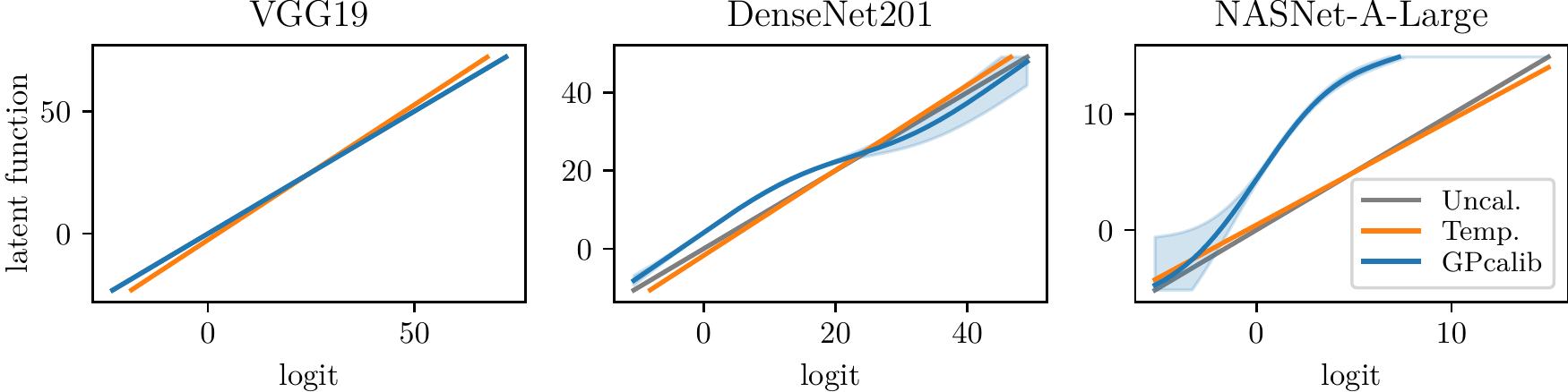}
	\end{subfigure}\\
	\vspace{1.5em}
	\begin{subfigure}[b]{0.9\textwidth}
		\includegraphics[width=\textwidth]{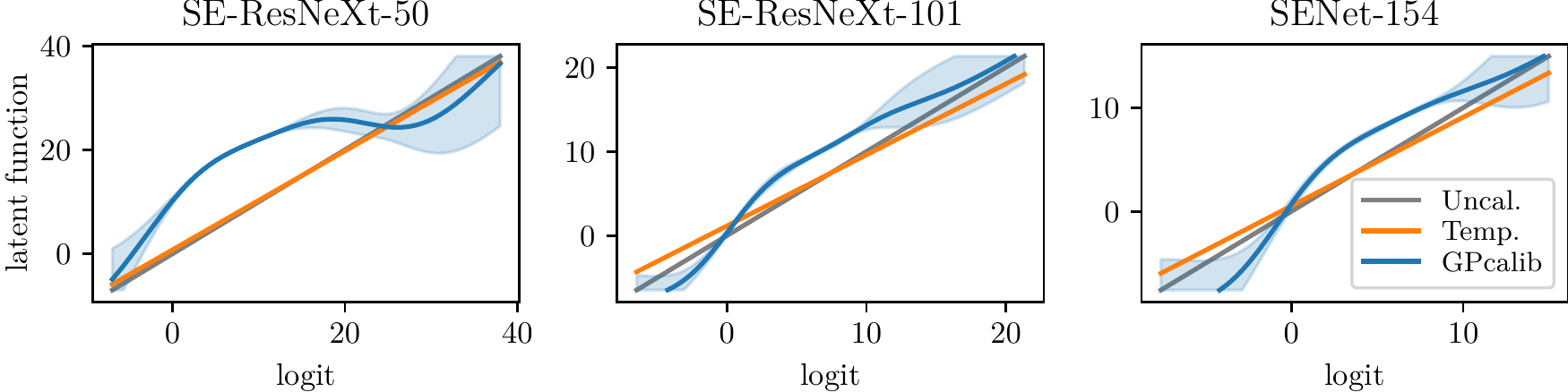}
	\end{subfigure}
	\caption{\textit{Illustration of non-linear latent functions in logit-space.} Additional latent maps of temperature scaling and GPcalib from our experiments on ImageNet in \Cref{sec:calibration_results} are shown. A higher degree of non-linearity and deviation from the identity map corresponds to a larger decrease in calibration error. Larger uncertainty of the latent Gaussian process corresponds to less samples in the calibration data set with logits in that range.\label{fig:latent_maps_logit_supp}}
\end{figure*}

In the multi-class calibration experiment on ImageNet, we observed better calibration of GPcalib for higher accuracy CNNs. In \Cref{fig:latent_maps_logit_supp}, we show additional latent functions obtained from the experiment. For VGG19 and DenseNet-201, the underlying network had comparably low calibration error to begin with. Both temperature scaling and GPcalib did not improve calibration significantly. This is reflected in latent space, where they do not deviate much from the identity map. However, in the case of SE-ResNeXt-50, SE-ResNeXt-101, SENet-154 and NASNet-A-Large GPcalib improved calibration noticably over the baseline and temperature scaling. Again, this improvement corresponds to a large change from the identity map in its latent function. 

\subsection{Reliability Diagrams}
\begin{figure*}[ht!]
    \centering
    \includegraphics[width=0.9\textwidth]{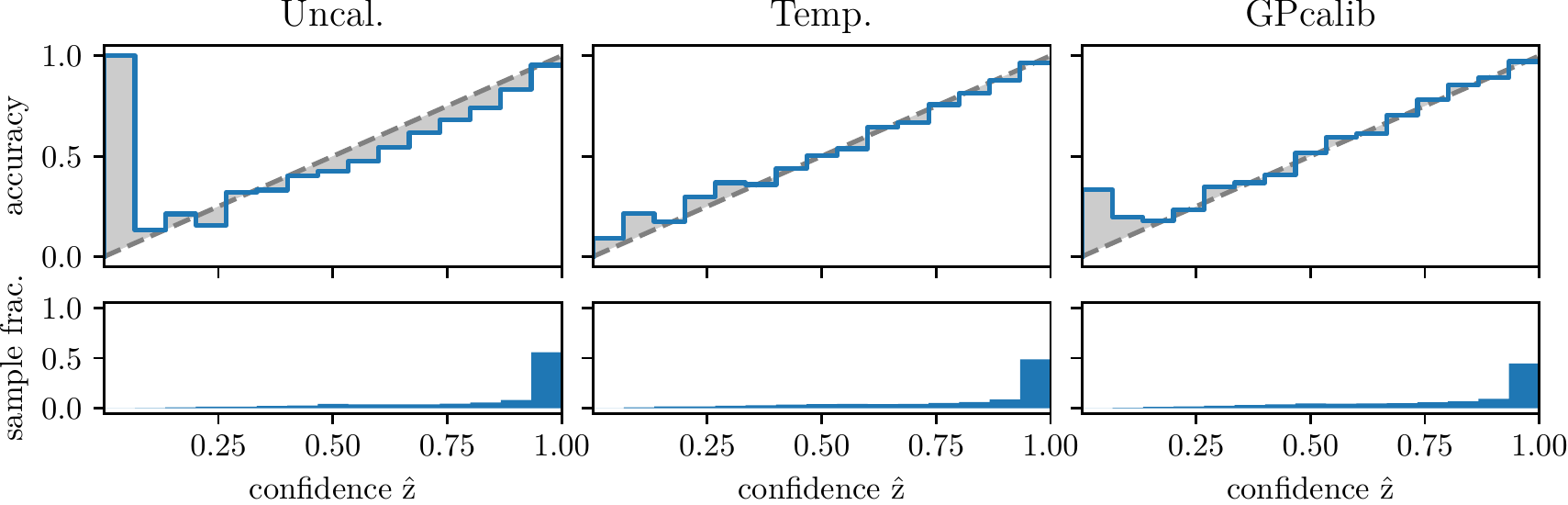}
    \caption{\textit{Reliability diagrams of ResNet-152.} ResNet-152 is overconfident for most of its output and thus miscalibrated. Note, that the large deviation for the left-most bin is an artifact of the low number of samples in that bin and thus not representative of the true accuracy for low confidence predictions. Both temperature scaling and GPcalib improve the classifier's calibration by shifting its confidence estimates closer to the actual accuracy.\label{fig:reldiagram_resnet152}}
\end{figure*}

\begin{figure*}[ht!]
    \centering
    \includegraphics[width=0.9\textwidth]{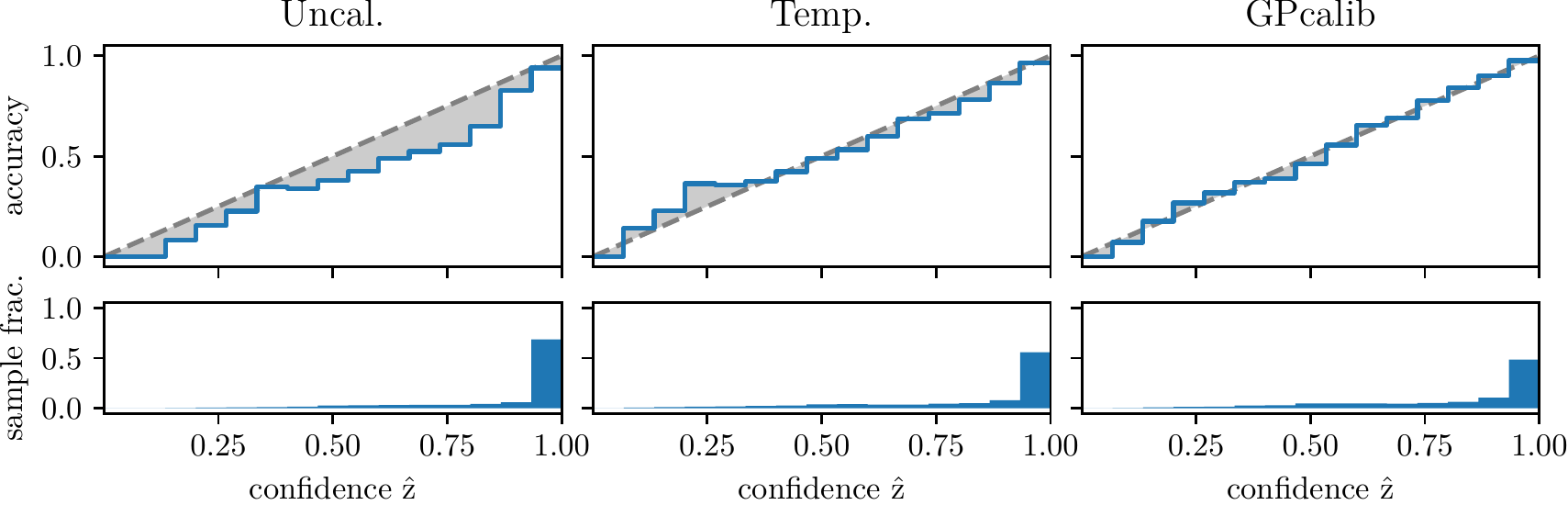}
    \caption{\textit{Reliability diagrams of PolyNet.} PolyNet shows a similar reliability curve as Resnet-152 in \Cref{fig:reldiagram_resnet152}. It also displays less accuracy per bin for the given confidence in its prediction. The shown calibration methods remedy this by reducing the confidence in some of its most certain estimates.\label{fig:reldiagram_polynet}}
\end{figure*}

\begin{figure*}[ht!]
    \centering
    \includegraphics[width=0.9\textwidth]{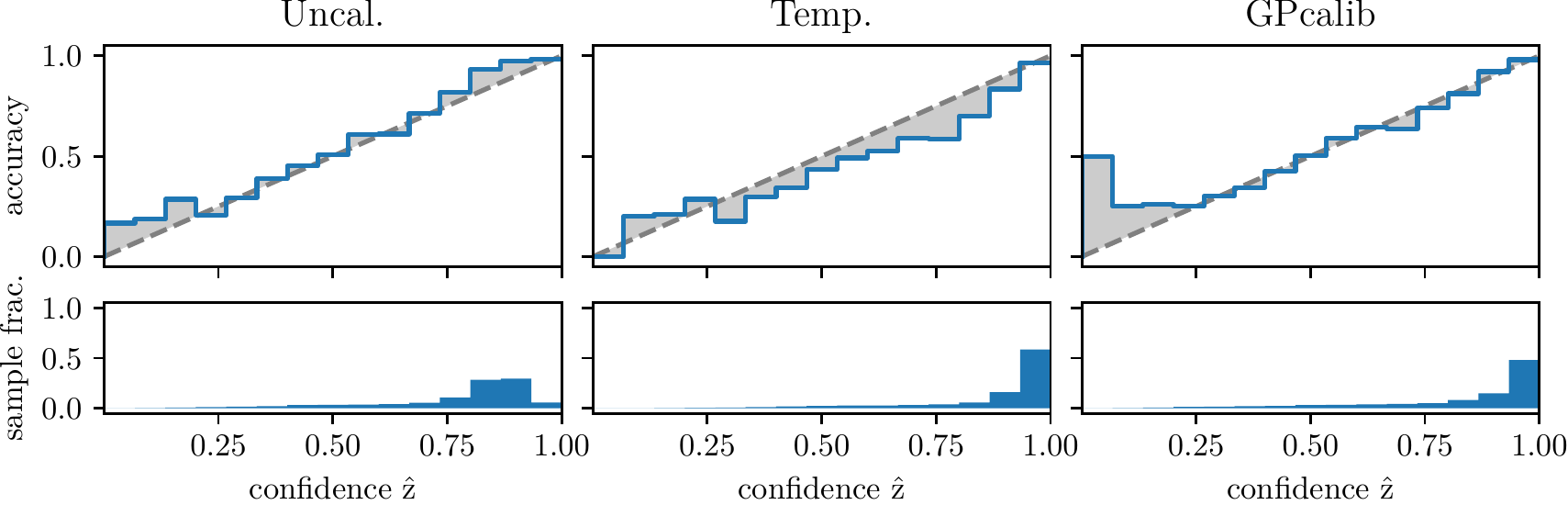}
\caption{\textit{Reliability diagrams of PNASNet-5-Large.} Contrary to most other CNNs, PNASNet-5-Large is actually less confident than accurate for much of its confidence histogram. We observed a similar phenomenon for other high accuracy CNNs on ImageNet in our experiments. Both calibration methods shift the histogram to more confident estimates, but GPcalib does not overcompensate for less confident predictions in contrast to temperature scaling.\label{fig:reldiagram_pnasnet5large}}
\end{figure*}

Reliability diagrams \citep{DeGroot1983, Niculescu-Mizil2005} visualize the degree of calibration of a model. They consist of a plot comparing confidence estimates with accuracy for a given binning of \([0,1]\) and a histogram of confidence estimates. They relate to the \(\operatorname{ECE}_1\) in the following way. The gray deviation from the diagonal in \Cref{fig:reldiagram_resnet152,fig:reldiagram_polynet,fig:reldiagram_pnasnet5large} weighted by the histogram below equals the estimate of the \(\operatorname{ECE}_1\) for a given binning. Here, for visualization purposes we chose 15 bins instead of 100 as in our experiments. For such a binning, most confidence estimates fall into one or two bins, leading to the estimation problem described in \Cref{sec:supp_number_of_bins}. When interpreting reliability diagrams keep in mind that bins with a low number of samples have high variance in their per-bin-accuracy. We show some reliability diagrams for CNNs on ImageNet from our experiments in \Cref{sec:calibration_results}. As is the case for most high-accuracy CNNs, Resnet-152 and PolyNet shown in \Cref{fig:reldiagram_resnet152,fig:reldiagram_polynet} mostly predict with very high confidence. Further, they generally predict higher confidence than accuracy, which is reflected in their high overconfidence. This observation is consistent with previous work on overconfidence of neural networks \citep{Lakshminarayanan2017, Guo2017, Hein2019}. Interestingly, PNASNet-5-Large (see \Cref{fig:reldiagram_pnasnet5large}) is actually underconfident. This also holds true for SENet-154 and NASNet-A-Large, which demonstrated the highest accuracy on ImageNet in our experiments.



\begin{sidewaystable}
  \caption{\textit{Accuracy of binary calibration experiments.} Average accuracy and standard deviation of 10 Monte-Carlo cross validation folds on binary benchmark data sets.}
  \label{tab:accuracy_binary_experiments}
  \centering
 {\small
\begin{filecontents*}{tables/binary_accuracy.csv}
Dataset,Model,uncalaccmean,uncalaccstd,plattaccmean,plattaccstd,isotonicaccmean,isotonicaccstd,betaaccmean,betaaccstd,bbqaccmean,bbqaccstd,tempaccmean,tempaccstd,gpcalibaccmean,gpcalibaccstd,gpcalibapproxaccmean,gpcalibapproxaccstd
KITTI,AdaBoost,.9463,.0006,.9499,.0009,.9497,.0006,.9499,.0009,.9444,.0072,.9463,.0006,.9465,.0005,.9463,.0006
KITTI,XGBoost,.9674,.0006,.9673,.0007,.9660,.0017,.9671,.0011,.9640,.0043,.9674,.0006,.9675,.0006,.9674,.0006
KITTI,Mondrian Forest,.9536,.0003,.9539,.0004,.9523,.0021,.9532,.0009,.9439,.0041,.9536,.0003,.9536,.0004,.9536,.0003
KITTI,Random Forest,.9639,.0007,.9628,.0011,.9616,.0017,.9625,.0017,.8922,.0055,.9639,.0007,.9637,.0007,.9639,.0007
KITTI,1 layer NN,.9620,.0006,.9644,.0007,.9686,.0012,.9684,.0009,.9647,.0060,.9620,.0006,.9620,.0006,.9620,.0006
PCam,AdaBoost,.7586,.0022,.7609,.0030,.7644,.0025,.7610,.0030,.7638,.0032,.7586,.0022,.7588,.0020,.7586,.0022
PCam,XGBoost,.8086,.0016,.8065,.0013,.8050,.0018,.8068,.0015,.8020,.0066,.8086,.0016,.8084,.0016,.8086,.0016
PCam,Mondrian Forest,.7946,.0012,.7976,.0013,.7954,.0032,.7976,.0017,.7950,.0027,.7946,.0012,.7946,.0013,.7946,.0012
PCam,Random Forest,.8487,.0007,.8484,.0015,.8473,.0016,.8482,.0016,.8110,.0041,.8487,.0007,.8483,.0008,.8487,.0007
PCam,1 layer NN,.5925,.0008,.6239,.0070,.6504,.0019,.6487,.0031,.6458,.0082,.5925,.0008,.5779,.0041,.5768,.0033
\end{filecontents*}
\csvreader[
	head to column names,
	tabular=llcccccccc,
	table head=\toprule Data Set & Model & Uncal. & Platt & Isotonic & Beta & BBQ & Temp. & GPcalib & GPcalib\\ &&&&&&&&&mean appr.\\ \midrule,
	table foot=\bottomrule,
	separator=comma]
	{tables/binary_accuracy.csv}
	{}
	{\Dataset & \Model & $\uncalaccmean$ & $\plattaccmean \pm \plattaccstd$ & $\isotonicaccmean \pm \isotonicaccstd$ & $\betaaccmean \pm \betaaccstd$ & $\bbqaccmean \pm \bbqaccstd$ & $\tempaccmean \pm \tempaccstd$ & $\gpcalibaccmean \pm \gpcalibaccstd$ & $\gpcalibapproxaccmean \pm \gpcalibapproxaccstd$}}
\end{sidewaystable}

\begin{sidewaystable}[t!]
  \caption{\textit{Binary calibration experiments.} Average \(\text{ECE}_1\) and standard deviation of 10 Monte-Carlo cross validation folds on binary benchmark data sets. Calibration errors (\(\text{ECE}_1\)) within one standard deviation of lowest per data set and model are printed in bold.}
  \label{tab:ece_binary_experiments}
  \centering
{\small
\begin{filecontents*}{tables/binary_ece.csv}
Dataset,Model,uncalECEmean,uncalECEstd,plattECEmean,plattECEstd,isotonicECEmean,isotonicECEstd,betaECEmean,betaECEstd,bbqECEmean,bbqECEstd,tempECEmean,tempECEstd,gpcalibECEmean,gpcalibECEstd,gpcalibapproxECEmean,gpcalibapproxECEstd
KITTI,AdaBoost,.4301,.0005,.0182,.0018,\textbf{.0134},.0021,.0180,.0016,.0190,.0055,.0185,.0017,.0192,.0018,.0187,.0014
KITTI,XGBoost,.0434,.0007,.0198,.0019,\textbf{.0114},.0026,.0178,.0015,.0184,.0038,.0204,.0009,.0186,.0017,.0181,.0018
KITTI,Mondrian Forest,.0546,.0006,.0198,.0011,\textbf{.0142},.0018,.0252,.0099,.0218,.0035,.0200,.0008,.0202,.0018,.0203,.0019
KITTI,Random Forest,.0768,.0006,.0147,.0027,\textbf{.0135},.0030,.0159,.0027,.0652,.0469,\textbf{.0126},.0020,.0182,.0032,\textbf{.0135},.0026
KITTI,1 layer NN,.0153,.0010,.0285,.0034,\textbf{.0121},.0043,.0174,.0026,.0178,.0056,.0280,.0015,\textbf{.0156},.0020,\textbf{.0157},.0020
PCam,AdaBoost,.2506,.0022,.0409,.0020,\textbf{.0335},.0047,\textbf{.0397},.0024,\textbf{.0330},.0077,\textbf{.0381},.0033,\textbf{.0389},.0032,\textbf{.0388},.0027
PCam,XGBoost,.0605,.0007,\textbf{.0378},.0010,\textbf{.0323},.0058,\textbf{.0356},.0028,\textbf{.0312},.0110,\textbf{.0399},.0020,\textbf{.0341},.0019,\textbf{.0354},.0026
PCam,Mondrian Forest,.0415,.0007,.0428,.0024,\textbf{.0291},.0066,\textbf{.0349},.0040,.0643,.0161,.0427,.0013,\textbf{.0352},.0040,\textbf{.0344},.0040
PCam,Random Forest,.0798,.0006,.0237,.0035,.0233,.0052,.0293,.0053,.0599,.0084,\textbf{.0210},.0013,.0283,.0027,\textbf{.0212},.0020
PCam,1 layer NN,.2090,.0009,.0717,.0051,\textbf{.0297},.0092,.0501,.0049,\textbf{.0296},.0102,.0542,.0015,.0454,.0031,.0487,.0032
\end{filecontents*}
\csvreader[
	head to column names,
	tabular=llcccccccc,
	table head=\toprule Data Set & Model & Uncal. & Platt & Isotonic & Beta & BBQ & Temp. & GPcalib & GPcalib\\ &&&&&&&&&mean appr.\\ \midrule,
	table foot=\bottomrule,
	separator=comma]
	{tables/binary_ece.csv}
	{}
	{\Dataset & \Model & $\uncalECEmean$ & $\plattECEmean \pm \plattECEstd$ & $\isotonicECEmean \pm \isotonicECEstd$ & $\betaECEmean \pm \betaECEstd$ & $\bbqECEmean \pm \bbqECEstd$ & $\tempECEmean \pm \tempECEstd$ & $\gpcalibECEmean \pm \gpcalibECEstd$ & $\gpcalibapproxECEmean \pm \gpcalibapproxECEstd$}}
\end{sidewaystable}

\begin{sidewaystable*}[ht!]
  \caption{\textit{Accuracy of multi-class calibration experiments.} Average accuracy and standard deviation of 10 Monte-Carlo cross validation folds on multi-class benchmark data sets.}
  \label{tab:accuracy_multiclass_experiments}
  \centering
 {\small
\begin{filecontents*}{tables/multiclass_accuracy.csv}
Dataset,Model,uncalaccmean,uncalaccstd,plattaccmean,plattaccstd,isotonicaccmean,isotonicaccstd,betaaccmean,betaaccstd,bbqaccmean,bbqaccstd,tempaccmean,tempaccstd,gpcalibaccmean,gpcalibaccstd,gpcalibapproxaccmean,gpcalibapproxaccstd
MNIST,AdaBoost,.7311,.0009,.6601,.0097,.6787,.0049,.6642,.0090,.6540,.0061,.7311,.0009,.7289,.0020,.7285,.0034
MNIST,XGBoost,.9333,.0006,.9330,.0011,.9312,.0014,.9331,.0011,.9274,.0022,.9333,.0006,.9333,.0006,.9333,.0006
MNIST,Mondrian Forest,.9133,.0008,.9144,.0015,.9118,.0014,.9142,.0015,.7475,.0138,.9133,.0008,.9132,.0008,.9152,.0009
MNIST,Random Forest,.9448,.0006,.9461,.0012,.9445,.0010,.9453,.0010,.0004,.0003,.9448,.0006,.9457,.0011,.9658,.0007
MNIST,1 layer NN,.9625,.0007,.9624,.0007,.9620,.0011,.9626,.0011,.9557,.0026,.9625,.0007,.9517,.0011,.9742,.0020
CIFAR-100,AlexNet,.4390,.0019,.4380,.0040,.4224,.0040,.4313,.0038,.3112,.0119,.4390,.0019,.4389,.0021,.4390,.0019
CIFAR-100,WideResNet,.8183,.0014,.8105,.0019,.8048,.0021,.8084,.0019,.7822,.0035,.8183,.0014,.8181,.0013,.8260,.0014
CIFAR-100,ResNeXt-29 (8x64),.8260,.0014,.8188,.0020,.8137,.0032,.8176,.0023,.7941,.0020,.8260,.0014,.8259,.0015,.8268,.0012
CIFAR-100,ResNeXt-29 (16x64),.8268,.0012,.8181,.0031,.8146,.0027,.8181,.0034,.7966,.0043,.8268,.0012,.8267,.0012,.8274,.0010
CIFAR-100,DenseNet-BC-190,.8274,.0010,.8218,.0018,.8155,.0021,.8189,.0019,.7938,.0040,.8274,.0010,.8275,.0010,.8183,.0014
ImageNet,AlexNet,.5652,.0040,.3475,.0071,.3455,.0056,.3528,.0058,.1866,.0050,.5652,.0040,.5653,.0040,.5652,.0040
ImageNet,VGG19,.7237,.0028,.4473,.0122,.4563,.0125,.4554,.0140,.2569,.0082,.7237,.0028,.7237,.0028,.7237,.0028
ImageNet,ResNet-50,.7614,.0042,.4710,.0087,.4810,.0099,.4790,.0097,.2660,.0071,.7614,.0042,.7614,.0041,.7614,.0042
ImageNet,ResNet-152,.7828,.0022,.4821,.0077,.4931,.0081,.4902,.0078,.2779,.0085,.7828,.0022,.7830,.0024,.7828,.0022
ImageNet,DenseNet-121,.7456,.0032,.4613,.0093,.4711,.0100,.4692,.0093,.2445,.0054,.7456,.0032,.7456,.0032,.7456,.0032
ImageNet,DenseNet-201,.7699,.0036,.4739,.0077,.4833,.0086,.4808,.0077,.2574,.0049,.7699,.0036,.7699,.0035,.7699,.0036
ImageNet,InceptionV4,.8011,.0048,.4958,.0121,.5061,.0118,.5043,.0117,.2575,.0056,.8011,.0048,.8011,.0047,.8011,.0048
ImageNet,SE-ResNeXt-50,.7905,.0056,.4849,.0097,.4965,.0097,.4945,.0100,.3136,.0102,.7905,.0056,.7904,.0056,.7905,.0056
ImageNet,SE-ResNeXt-101,.8022,.0042,.4995,.0118,.5087,.0121,.5081,.0121,.2480,.0058,.8022,.0042,.8021,.0043,.8022,.0042
ImageNet,PolyNet,.8086,.0030,.4987,.0082,.5122,.0099,.5090,.0098,.3232,.0096,.8086,.0030,.8083,.0030,.8086,.0030
ImageNet,SENet-154,.8117,.0030,.5035,.0070,.5138,.0061,.5144,.0062,.1943,.0055,.8117,.0030,.8117,.0030,.8117,.0030
ImageNet,PNASNet-5-Large,.8289,.0041,.5146,.0119,.5239,.0125,.5242,.0123,.1951,.0085,.8289,.0041,.8290,.0041,.8289,.0041
ImageNet,NASNet-A-Large,.8234,.0026,.5135,.0099,.5239,.0105,.5235,.0103,.2009,.0103,.8234,.0026,.8233,.0027,.8234,.0026
\end{filecontents*}
\csvreader[
	head to column names,
	tabular=llcccccccc,
	table head=\toprule &&& \multicolumn{4}{c}{one-vs-all}\\ \cmidrule(r){4-7}Data Set & Model & Uncal. & Platt & Isotonic & Beta & BBQ & Temp. & GPcalib & GPcalib\\ &&&&&&&&&mean appr.\\ \midrule,
	table foot=\bottomrule,
	separator=comma]
	{tables/multiclass_accuracy.csv}
	{}
	{\Dataset & \Model & $\uncalaccmean$ & $\plattaccmean \pm \plattaccstd$ & $\isotonicaccmean \pm \isotonicaccstd$ & $\betaaccmean \pm \betaaccstd$ & $\bbqaccmean \pm \bbqaccstd$ & $\tempaccmean \pm \tempaccstd$ & $\gpcalibaccmean \pm \gpcalibaccstd$ & $\gpcalibapproxaccmean \pm \gpcalibapproxaccstd$}}
\end{sidewaystable*}

\begin{sidewaystable*}[t!]
  \caption{\textit{Multi-class calibration experiments.} Average \(\text{ECE}_1\) and standard deviation of 10 Monte-Carlo cross validation folds on multi-class benchmark data sets. Calibration errors (\(\text{ECE}_1\)) within one standard deviation of lowest per data set and model are printed in bold.}
  \label{tab:ece_std_multiclass_experiments}
  \centering
{\small
\begin{filecontents*}{tables/multiclass_ece_gpcalibapprox.csv}
Dataset,Model,uncalECEmean,uncalECEstd,plattECEmean,plattECEstd,isotonicECEmean,isotonicECEstd,betaECEmean,betaECEstd,bbqECEmean,bbqECEstd,tempECEmean,tempECEstd,gpcalibECEmean,gpcalibECEstd,gpcalibapproxECEmean,gpcalibapproxECEstd
MNIST,AdaBoost,.6121,.0009,.2267,.0137,.1319,.0108,.2222,.0134,.1384,.0104,.1567,.0122,\textbf{.0414},.0085,\textbf{.0428},.0123
MNIST,XGBoost,.0740,.0005,.0449,.0021,\textbf{.0176},.0018,\textbf{.0184},.0014,.0207,.0020,.0222,.0015,\textbf{.0180},.0014,\textbf{.0182},.0016
MNIST,Mondrian Forest,.2163,.0008,.0357,.0049,.0282,.0021,.0383,.0057,.0762,.0111,\textbf{.0208},.0012,\textbf{.0213},.0020,\textbf{.0206},.0019
MNIST,Random Forest,.1178,.0008,.0273,.0039,.0207,.0042,.0259,.0070,.1233,.0005,\textbf{.0121},.0012,.0148,.0021,.0135,.0009
MNIST,1 layer NN,.0262,.0005,\textbf{.0126},.0031,.0140,.0017,.0168,.0018,.0186,.0027,.0195,.0060,.0239,.0023,\textbf{.0109},.0017
CIFAR-100,AlexNet,.2751,.0016,.0720,.0090,.1232,.0102,.0784,.0091,.0478,.0058,\textbf{.0365},.0024,\textbf{.0369},.0054,\textbf{.0377},.0049
CIFAR-100,WideResNet,.0664,.0009,.0838,.0056,.0661,.0040,.0539,.0031,.0384,.0046,.0444,.0030,.0283,.0027,\textbf{.0246},.0027
CIFAR-100,ResNeXt-29 (8x64),.0495,.0008,.0882,.0040,.0599,.0062,.0492,.0054,.0392,.0088,.0424,.0015,\textbf{.0251},.0020,\textbf{.0266},.0019
CIFAR-100,ResNeXt-29 (16x64),.0527,.0011,.0900,.0041,.0620,.0058,.0520,.0054,.0365,.0078,.0465,.0026,.0266,.0018,\textbf{.0241},.0021
CIFAR-100,DenseNet-BC-190,.0717,.0006,.0801,.0061,.0665,.0036,.0543,.0036,.0376,.0049,.0377,.0019,\textbf{.0237},.0021,.0273,.0025
ImageNet,AlexNet,\textbf{.0353},.0025,.1132,.0046,.2937,.0094,.2290,.0078,.1307,.0093,\textbf{.0342},.0035,\textbf{.0357},.0025,\textbf{.0356},.0024
ImageNet,VGG19,.0377,.0022,.0965,.0095,.2810,.0131,.2416,.0163,.1617,.0110,\textbf{.0342},.0034,\textbf{.0364},.0018,\textbf{.0362},.0016
ImageNet,ResNet-50,.0441,.0025,.0875,.0072,.2724,.0188,.2250,.0122,.1635,.0063,\textbf{.0341},.0018,\textbf{.0335},.0054,\textbf{.0330},.0055
ImageNet,ResNet-152,.0545,.0016,.0879,.0066,.2761,.0348,.2201,.0130,.1675,.0064,.0323,.0019,\textbf{.0283},.0038,\textbf{.0283},.0032
ImageNet,DenseNet-121,.0380,.0015,.0949,.0061,.2682,.0069,.2297,.0139,.1512,.0099,\textbf{.0329},.0020,.0357,.0040,.0360,.0033
ImageNet,DenseNet-201,.0410,.0032,.0898,.0111,.2706,.0210,.2189,.0157,.1614,.0116,\textbf{.0324},.0018,.0367,.0058,.0362,.0063
ImageNet,InceptionV4,.0318,.0025,.0865,.0102,.2900,.0351,.1653,.0138,.1593,.0155,.0462,.0070,\textbf{.0269},.0031,\textbf{.0274},.0027
ImageNet,SE-ResNeXt-50,.0440,.0033,.0889,.0101,.2684,.0260,.1789,.0106,.1990,.0176,.0482,.0069,\textbf{.0279},.0055,\textbf{.0280},.0045
ImageNet,SE-ResNeXt-101,.0574,.0026,.0853,.0112,.2844,.0218,.1631,.0098,.1496,.0106,.0415,.0039,\textbf{.0250},.0020,\textbf{.0253},.0017
ImageNet,PolyNet,.0823,.0031,.0806,.0048,.2590,.0302,.2006,.0124,.1787,.0151,.0369,.0030,\textbf{.0283},.0041,\textbf{.0276},.0027
ImageNet,SENet-154,.0612,.0015,.0809,.0106,.3003,.0349,.1582,.0098,.1502,.0127,.0497,.0028,\textbf{.0309},.0030,\textbf{.0303},.0030
ImageNet,PNASNet-5-Large,.0702,.0025,.0796,.0081,.3063,.0221,.1430,.0085,.1355,.0101,.0486,.0031,\textbf{.0270},.0018,\textbf{.0266},.0019
ImageNet,NASNet-A-Large,.0530,.0027,.0826,.0082,.3265,.0300,.1437,.0085,.1268,.0097,.0516,.0029,\textbf{.0255},.0030,\textbf{.0251},.0026
\end{filecontents*}
\csvreader[
	head to column names,
	tabular=llcccccccc,
	table head=\toprule &&& \multicolumn{4}{c}{one-vs-all}\\ \cmidrule(r){4-7}Data Set & Model & Uncal. & Platt & Isotonic & Beta & BBQ & Temp. & GPcalib & GPcalib\\ &&&&&&&&&mean appr.\\ \midrule,
	table foot=\bottomrule,
	separator=comma]
	{tables/multiclass_ece_gpcalibapprox.csv}
	{}
	{\Dataset & \Model & $\uncalECEmean$ & $\plattECEmean \pm \plattECEstd$ & $\isotonicECEmean \pm \isotonicECEstd$ & $\betaECEmean \pm \betaECEstd$ & $\bbqECEmean \pm \bbqECEstd$ & $\tempECEmean \pm \tempECEstd$ & $\gpcalibECEmean \pm \gpcalibECEstd$ & $\gpcalibapproxECEmean \pm \gpcalibapproxECEstd$}}
\end{sidewaystable*}

\end{document}